\pdfoutput=1

\documentclass[11pt]{article}
\usepackage{graphicx}
\usepackage{latexsym}
\usepackage{times}
\usepackage{soul}
\usepackage{url}
\usepackage[utf8]{inputenc}
\usepackage{graphicx}
\usepackage{amsmath}
\usepackage{amsthm}
\usepackage{booktabs}
\usepackage{algorithm}
\usepackage{algorithmic}
\usepackage{multirow}
\usepackage{amssymb}
\usepackage{xcolor}

\usepackage{hyperref}
 
\usepackage{epstopdf}
\usepackage{tikz-cd}
\usepackage[]{EMNLP2023}

\usepackage{times}
\usepackage{latexsym}

\usepackage[T1]{fontenc}

\usepackage[utf8]{inputenc}

\usepackage{microtype}

\usepackage{inconsolata}

\title{SemRoDe: Macro Adversarial Training to Learn Representations That are Robust to Word-Level Attacks}

\author{Brian Formento$^{1,2}$, Wenjie Feng$^{1}$, Chuan Sheng Foo$^{2,3}$, Luu Anh Tuan$^{4}$, See-Kiong Ng$^{1}$ \\  
Institute of Data Science, National University of Singapore$^1$\\ Institute for Infocomm Research, A*STAR$^2$ \\ Centre for Frontier AI Research, A*STAR$^3$,  Nanyang Technological University$^4$ \\
\texttt{brian.formento@u.nus.edu},  \texttt{wenjie.feng@nus.edu.sg}, \texttt{foo\_chuan\_sheng@i2r.astar.edu.sg}
}
\begin{document}
\maketitle
\begin{abstract}
Language models (LMs) are indispensable tools for natural language processing tasks, but their vulnerability to adversarial attacks remains a concern. While current research has explored adversarial training techniques, their improvements to defend against word-level attacks have been limited. In this work, we propose a novel approach called Semantic Robust Defence (SemRoDe), a Macro Adversarial Training strategy to enhance the robustness of LMs. Drawing inspiration from recent studies in the image domain, we investigate and later confirm that in a discrete data setting such as language, adversarial samples generated via word substitutions do indeed belong to an adversarial domain exhibiting a high Wasserstein distance from the base domain. Our method learns a robust representation that bridges these two domains. We hypothesize that if samples were not projected into an adversarial domain, but instead to a domain with minimal shift, it would improve attack robustness. We align the domains by incorporating a new distance-based objective. With this, our model is able to learn more generalized representations by aligning the model's high-level output features and therefore better handling unseen adversarial samples. This method can be generalized across word embeddings, even when they share minimal overlap at both vocabulary and word-substitution levels. To evaluate the effectiveness of our approach, we conduct experiments on BERT and RoBERTa models on three datasets. The results demonstrate promising state-of-the-art robustness.

\end{abstract}

\section{Introduction}

Basic deep learning models are not inherently robust. This issue has been illustrated in numerous studies pointing to structural problems such as dataset shift \cite{Datashift} and the prevalence of adversarial attacks \cite{Attacksurvey}. Particularly in the context of natural language research, adversarial attacks are generated by introducing small perturbations either at the character, word, or sentence level of a textual input while maintaining the meaning, semantics, and grammar of the sentence. The development of robust systems is therefore of paramount importance for a multitude of reasons. Notably, content filters in need of detecting offensive language may be misled into classifying negative content as positive. Furthermore, in the realm of language model learning, adversarial attacks can successfully circumvent content moderation pipelines, prompting harmful content generation which can severely impact individuals and society.


With the accelerating popularity of LLMs and applications, there has been an urgency for robustness research.

There is a small, underexplored body of work that has proposed the merger of an adversarial distribution with the original distribution to improve a model's adversarial robustness. Preliminary and promising results have emerged when implementing a theoretical solution based on base-adversarial alignment in the form of domain adaptation in the image domain \cite{AT_Domain_Adaptation,OT_Images}.




Inspired by these findings, we initially expand on their hypothesis, demonstrating how a domain adaptation solution may have limitations. However, we maintain the central idea that aligning the base and adversarial features can lead to robust results and explore the validity of this theory in the discrete language domain.

We further refine this approach by formulating a solution that aligns more closely with previous work in adversarial robustness, making it more relevant to the language domain. Through this process, we demonstrate state-of-the-art performance across a variety of models and datasets, thereby advancing the cause of improving adversarial robustness in the discrete language domain.

The following points summarize the major contributions of our research:

\begin{itemize}
 \item We demonstrate that adversarial samples introducing word substitutions \cite{A2T,Textfooler} lead to unwanted distributions in feature space (Figure \ref{img:aug_data_distribution}), resulting in a high Wasserstein distance between base and adversarial features (Figure \ref{img:distance_comparison_plot}). We propose a solution through a regularizer that reduces the distance between a base and adversarial domain, aiding the formulating of a representation robust to potential attacker projections.

\item We explore the function of distance-based regularizers in aligning the base and adversarial language domains. The effectiveness of feature alignment in enhancing robustness is evidenced in our distribution-oriented modelling approach, which shows strong generalization against various attack algorithms.

\item Lastly, our work shows competitive performance across multiple datasets and models, offering preliminary findings on the usage of diverse distance regularizers to accomplish distribution alignment. This includes regularizers such as Maximum Mean Discrepancy (MMD), CORrelation ALignment (CORAL), and Optimal Transport. Our code is publicly available\footnote{\href{https://github.com/Aniloid2/SemRoDe-MacroAdversarialTraining}{https://github.com/Aniloid2/SemRoDe-MacroAdversarialTraining}}.

\end{itemize}

\section{Related Work}

\paragraph{Attacks on NLP systems}
 
On the front of adversarial attacks, recent work has developed methods to attack NLP systems. These methods firstly generate substitution candidates by removing, swapping, or adding letters/punctuation \cite{Hotflip,Hero,Viper,SSTA} or finding word candidates around the $\ell_2$ \cite{L2ball1,L2ball2} or convex hull \cite{Convexhull} of the word embedding space. These word candidates can be general word substitutions \cite{TextBugger}, synonyms \cite{Textfooler, BERTAttack, PWWS}, sememes \cite{Sememepso}, or grammatical inflections \cite{Inflection}. The substitution candidate after that is expected to preserve a certain level of semantic structure, which is often enforced with the Levenshtein distance \cite{Deepwordbug} or through the use of a semantic encoder, \cite{Universalsentenceencoder,Bertsentence}. These candidates can then be applied to the base sentence in a black-box, gray-box, or white-box setting. The black-box setting uses no information from the victim's system. In the gray-box settings, some information from the model is used, such as the output logit or label. In the white-box setting, all model information is available, including weights. In the case of the gray-box setting, such character or word changes can result in a successful perturbation when performing convex optimization over the input by replacing the original letter or word with the potential change and analyzing the logit or label. This optimization problem is often solved using greedy search \cite{Textfooler}, with genetic algorithms \cite{Genetic}, or particle swarm optimization \cite{Sememepso}.  On the contrary, white-box attacks leverage gradient and parameters information from the model to inject a $\delta$ in the embedding space, rather than the input space. 
 
\paragraph{Defence methods}
Research into the generation of adversarial samples can subsequently contribute to enhancing model robustness through adversarial training, early efforts in adversarial training in NLP included the adversarial examples back into training by including the adversarial samples in the batch \cite{Textfooler}. More recently, these strategies have been refined \cite{A2T}. Alternatively, it is possible to extend the original objective by adding an adversarial regularizer, similar to how adversarial training is performed in the image domain \cite{AT,AT2,JAEAT,CreAT}. Building upon this line of work, embedding perturbation methods such as FreeLB \cite{FreeLB}, ASCC \cite{Convexhull}, InfoBERT \cite{InfoBERT}, DSRM \cite{DSRM} and FreeLB++ \cite{TextDefender}, leverage the model's gradient to apply a perturbation within the continuous embedding space via projected gradient descent. This perturbation is constrained by a specified norm, such as the L2 norm \cite{L2ball1,L2ball2}, or maintained within the confines of a convex hull \cite{Convexhull}. Additionally, some approaches only induce perturbation when the loss falls below a predefined threshold \cite{FloodingX}. These adversarial training schemas result in a min-max game between the base and adversarial objectives, forcing the model to learn both datasets. 

More recently, defense techniques originating from the off-manifold conjecture in images have been adapted for the NLP context \cite{TMD}. Additionally, a detector \cite{Detector} and an encoder \cite{Semantic_Encoder} for adversarial samples have been proposed. On the other hand, it is also feasible to train an anomaly detector to identify word substitutions \cite{Anomaly_Detection} or adversarial sentences \cite{TextShield}. The primary drawback of these methods is the need to incorporate extra components into the pipeline, either before or after inference, resulting in additional computational load. A different line of work demonstrated that enhancing the robustness of attention-based models to adversarial attacks can be achieved by aligning the distributions of the keys and queries \cite{Alignment_Attention}. Adversarial training can be effectively utilized to defend against word-level attacks without introducing extra computational overhead. There are also several works exploring certified robustness for NLP systems but these methods remain hard to scale to large models \cite{IBP,SAFER,Confounder_Certified,Robustness_Aware}.
A comprehensive benchmarking of adversarial defense methods \cite{TextDefender} concluded that FreeLB++ performs the best; it employs embedding perturbations to acquire robust invariant representations. FreeLB++ is the state-of-the-art technique, together with DSRM to improve general robustness, while we identify A2T as being the state-of-the-art technique to improve robustness through adversarial data augmentation. FreeLB++, is confined to exploring the embedding space, while A2T is able to explore the full semantic space.

\section{Background}
\subsection{Overview}
We consider the setting of a sequence classifier, composed of a base model $f^{base}_{\theta}$ whose output is pooled $f^{pool}_{\theta}$ and then fed into a classifier head $f^{cls}_{\theta}$ resulting in $f_{\theta}: t(\mathcal{X}) \mapsto \mathcal{Y}$, that takes an input embedding \textbf{X} $\leftarrow t(x)$. Where the input sequence of words $x = (\tau_1, \ldots, \tau_n) \in \mathcal{X}$ with the ground truth label $y$ and outputs a prediction $\hat{y} = f_{\theta}(t(x))$ are first parsed through tokenizer $t$. An adversarial attack with algorithm $g$ on input $x$ and classifier $f_{\theta}$ would perturb $\tau$, for example, using character manipulations or word substitutions, to produce a new adversarial sample $\hat{x}$ that is misclassified by $f_{\theta}$ such that $f_{\theta}(t(\hat{x})) \neq y$, the sample, can then be validated for linguistic characteristics with $d$.

\subsection{Adversarial training}
The objective of adversarial training in NLP is to minimize the high-risk regions in the continuous space around an input sample's word embeddings, while constraining the search area with a bound \cite{FreeLB}. Adversarial training is formally defined as follows:

\begin{equation}
\min_{\theta}\mathbb{E}_{\{x,y\}\sim D} \max_{ ||\delta|| 	\le \epsilon}\mathcal{L}(f_{\theta}(\textbf{X}+\delta), y)
\end{equation}

Here, $D$ represents the data distribution, $\textbf{X}$ represents a tensor of word embeddings, where $\textbf{X}$ is the embedding representation of all input tokens, and $\delta$ represents a perturbation. The aim of adversarial training is to minimize the regions of high risk around embedding $\textbf{X}$ with a bound delta \cite{FreeLB}. The constraint is usually the $\ell_2$ norm around the embedding. However, previous research has explored other bounds, such as the convex hull \cite{Convexhull}.

\subsection{Base domain}

In adversarial training the original, unperturbed samples $\mathcal{X} \mapsto \mathcal{Y}$ originate from a base domain. We expand this notion to distinguish between the base/adversarial domain in the input/linguistic space and the feature space. Firstly $
\mathcal{X}_{B} \triangleq \{\mathbf{x} \mid \mathbf{x} \text{ is a sample from base input domain}\}$ then $H^{B}_{\textbf{X}} \triangleq f^{pool}_{\theta}(t(\mathcal{X}_{B})) = \{f^{pool}_{\theta}(t(x)) \mid x \in \mathcal{X}_{B}\}$ where $H^{B}_{\textbf{X}}$ is the feature distribution of the samples from the base domain. Assume, inline with previous work \cite{AT_Domain_Adaptation}, the base domain $\mathcal{X}_{B}$ follows a multivariate normal distribution $\mathcal{X}_{B} \sim \mathcal{N}(\mu_{\mathcal{X}_{B}},\Sigma_{\mathcal{X}_{B}})$

\subsection{Adversarial domain}
A new perturbed sample $\hat{x}$ is obtained by adding a perturbation $\delta$ with strength controlled by maximize value $\epsilon$ to the original sample $x$,  i.e., $\hat{x} = x + \delta$  and $|\delta| \le \epsilon$.  To obtain more real perturbation samples for NLP task, we control the perturbation strength through a controller $g$, which is an abstraction of an attack algorithm, such as TextFooler, and perform the verbatim perturbation. Our assumption is that the NLP classification model works, as it  captures the semantics of the data and uses that to make a decision.

\begin{figure}[thp]
    \centering
    \includegraphics[width=.95\linewidth]{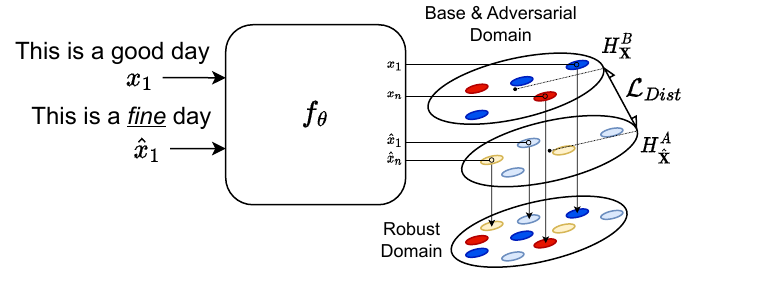} 
     
    \caption{ The statistical components are aligned in both the base and adversarial domain through the regularizer $\mathcal{L}_{Dist}$. Over time. This alignment allows the model $f$ to project both base and adversarial samples to a robust domain, thus enhancing the robust generalization to adversarial samples. 
    }\label{img:explenation}
\end{figure}

As we do an adversarial perturbation on sample $x$, this perturbation will introduce a covariate shift to the original sample, dataset or distribution. This will make the distribution bimodal on one class instead of the expected single smooth distribution as illustrated in Figure \ref{img:explenation} where a perturbation in the input leads to the sample being projected onto a different domain $H^{A}_{\hat{\textbf{X}}}$. Where $H^{A}_{\hat{\textbf{X}}} $, at times also denoted as $ H^{B}_{\textbf{X} + \delta}$, is the feature distribution of the samples from the adversarial domain. In our setup, after these samples have been passed through $f^{pool}_{\theta}$, they will be projected in an adversarial domain. Therefore, in our setup, $\mathcal{X}_{A} \triangleq \{\hat{x} \mid \hat{x} \text{ is a sample from adversarial input domain}\}$ and $H^{A}_{\hat{\textbf{X}}} \triangleq f^{pool}_{\theta}(t(\mathcal{X}_{A})) = \{f^{pool}_{\theta}(t(\hat{x})) \mid \hat{x} \in \mathcal{X}_{A}\}$. Assume the adversarial domain $\mathcal{X}_{A}$ follows a multivariate normal distribution $\mathcal{X}_{A} \sim \mathcal{N}(\mu_{\mathcal{X}_{A}},\Sigma_{\mathcal{X}_{A}})$

\begin{figure}[th]
    \centering
    \includegraphics[width=.68\linewidth]{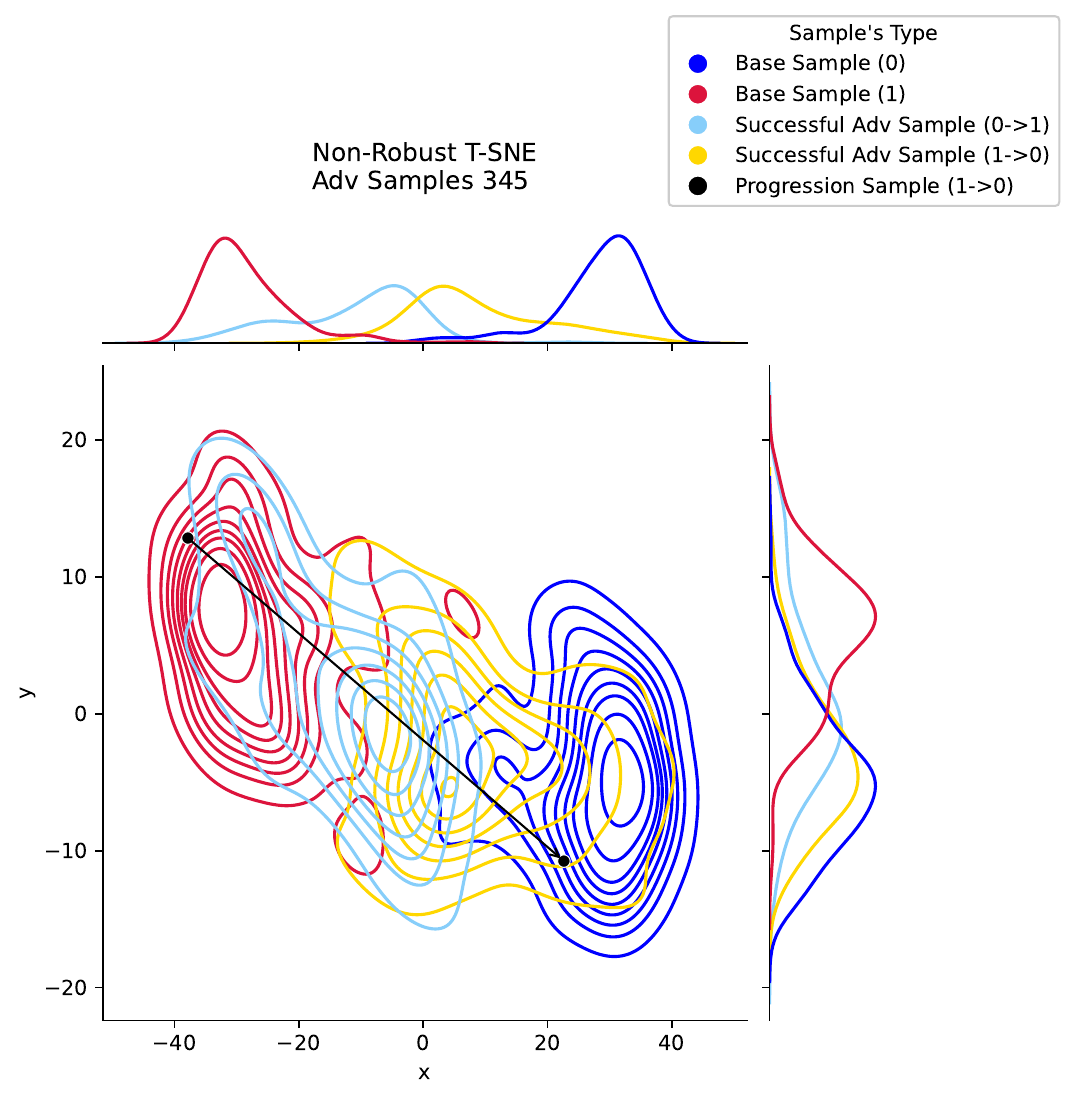} 
    \includegraphics[width=.68\linewidth]{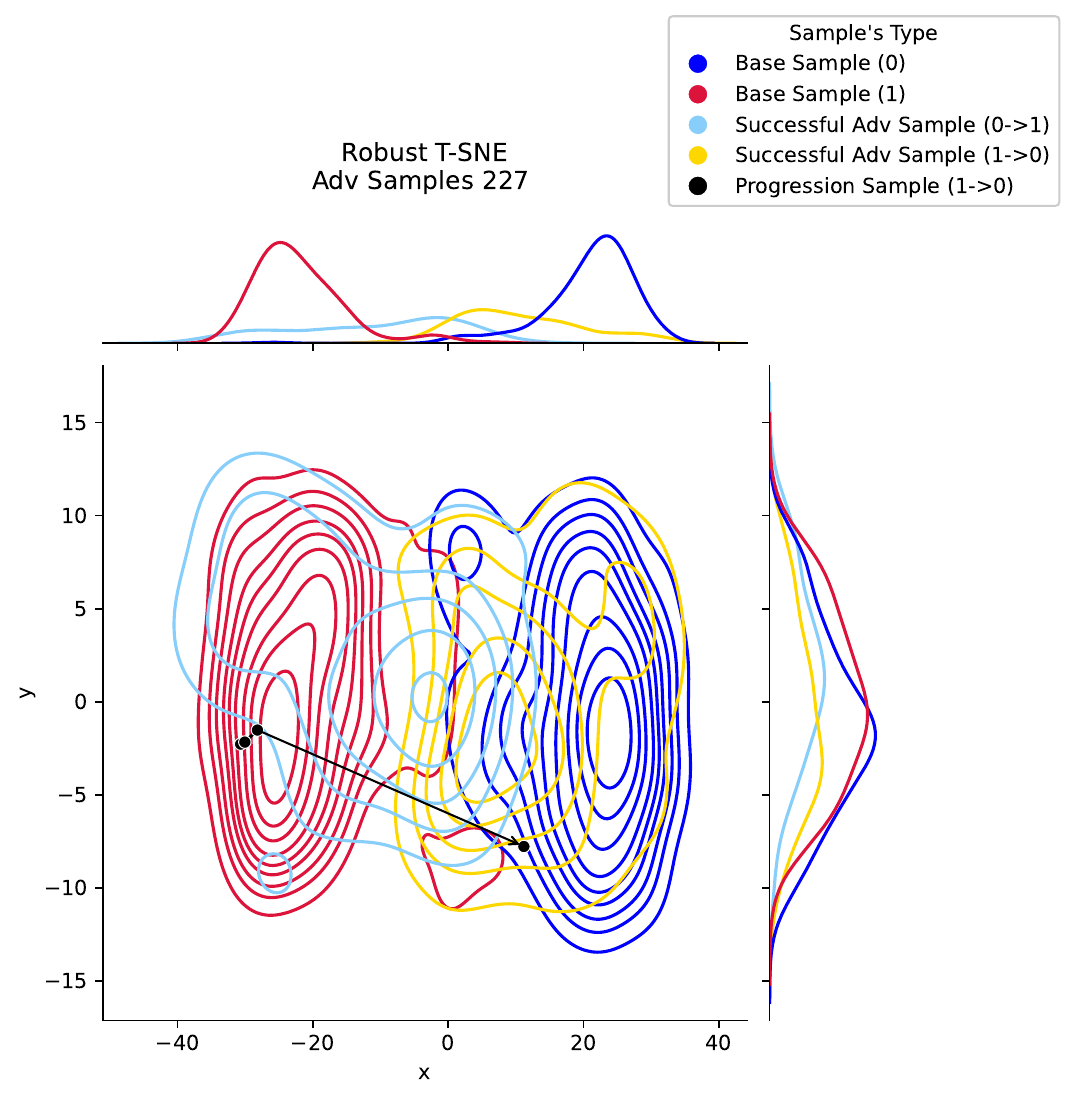} 
    \caption{
        \textbf{The distributions of the MR training dataset with t-SNE projection in a binary classification task.}
        Heavy overlapping (Top) of the augmented two-class data 
        leads to mixtures of marginal distributions, 
        which is alleviated and nearly linearly separable  (Bottom) after applying alignment between the original and augmentation distributions.
    }
    \label{img:aug_data_distribution}
\end{figure}

\paragraph{Domain alignment methods} 
Paraphrased samples, generated by substituting words, result in a significant distributional shift in the high-level representation space within the later layers of the model. Therefore, the distances between the two distributions at the $f^{pool}_{\theta}$ layer will be high, as shown in Figure \ref{img:distance_comparison_plot}. When t-SNE is applied to the output features after $f^{pool}_{\theta}$, there are $k$ clusters, half of which originate from the base samples while the other half from the adversarial samples (Figure \ref{img:aug_data_distribution}). The objective is thus to minimize these mean and covariance distances of the high-level features, so that only $k/2$ clusters remain, this would mean the adversarial and base samples are perfectly aligned in the feature space. The aim is to prevent adversarial samples from being projected within the adversarial feature distribution created by the model, and instead, have them projected onto a new robust feature distribution. To achieve this, we explore approaches to align feature spaces, ensuring that features from adversarial samples more accurately represent features from base samples.

     

Commonly-used methods for measuring distance in the feature space and perform feature alignment include Maximum Mean Discrepancy (MMD), Optimal Transport (OT), and CORrelation ALignment (CORAL). MMD measures distributional differences through the mean of feature representations, OT calculates the optimal transformation cost, and CORAL aligns second-order statistics. MMD is straightforward but sensitive to kernel choices. OT captures fine-grained differences but can be computationally demanding. CORAL is simple and efficient but may not capture all aspects of distributional discrepancy.

\section{Methodology}
Embedding robustness, as described by \cite{FreeLB}, is a concept whereby a model is considered robust if it can maintain low risk within regions encircling a specific embedding \textbf{X}. Adversarial training and gradient ascent methodologies are applied directly to word embedding by prominent techniques such as \cite{FreeLB, PGD}, in an effort to accomplish this. However, these methods display limitations in sustaining this robustness. These restrictions occur due to the model being bound by $\epsilon$. Transgression beyond this $\epsilon$ limit incites undesirable word substitutions (Table \ref{Qualitative_sample}). 


Notably, this presents a hurdle, since the majority of word-level attacks, not having access to the input word embeddings or gradients, conduct heuristic optimization on the input sample. Each word often has $N$ substitution candidates. Therefore, they employ word substitutions to introduce a $\delta$ into a sentence (see Figure \ref{img:large_delta_explenation} in the Appendix), thus often exceeding the $\epsilon$ restriction. This allows such attacks to explore regions in the semantic space that the victim's model has not previously seen.






Consequently, our research involves the direct utilization of existing attack algorithms to generate the adversarial data intended for training. Subsequently, we scrutinize the extent to which our training approach shows generalization when faced with different attack algorithms employed by an attacker.

\subsection{Perturbations}
In our work, we address the saddle problem in the following manner. Firstly, we adopt a dynamic $\delta$ that is tailored for a specific word substitution pair. This delta is determined by a word substitution algorithm, denoted as $g$. The sample is preserved only if it complies with quantitative quality standards set by $d$. Rather than maximizing the loss, we minimize the classification confidence. This strategy aligns better with the adversarial objective commonly employed in most NLP adversarial attacks. The process of creating adversarial samples with $g$ and minimizing the classification confidence inherently maximizes the distance between the base and adversarial domain. We theorize that this divergence is significantly larger than in the image domain, or when introducing small deltas in the embedding space, mainly because word substitutions contribute to a large $\delta$. This domain distance can be summarized by computing a loss value with distance-measuring techniques like MMD or CORAL.  (See Appendix \ref{Implementation_details}, \ref{large_delta_explenation} for further details)

\subsection{Perturbation alignment}
\label{section:Perturbation_alignment}

Through empirical analysis, we found that multiple perturbed samples tend to construct new distributions in the feature space. This presents an issue as the classifier component, modelled on the original distributions, continues to function based on them. This phenomenon is depicted in Figure \ref{img:aug_data_distribution} (top and bottom), where base samples are characterized by blue and red distributions, and adversarial samples by cyan and yellow distributions. We explain this phenomenon by the presence of a hypothetical base and adversarial distribution over the feature space ($H^{B}_{\hat{\textbf{X}}} $ and $H^{A}_{\hat{\textbf{X}}} $) which we give empirical evidence of existing through the Wasserstein distance in Figure \ref{img:distance_comparison_plot}. Our aim is to employ a regularizer during training that brings the distribution closer and hereby aligns the domains into a robust domain (Figure \ref{img:explenation}). This will enable the classifier to effectively function on both the adversarial (which will no longer be adversarial) and base samples. As shown in Figure \ref{img:aug_data_distribution} (bottom), this is demonstrated by the smoother distributions and a reduction in the number of samples in both the cyan and yellow distributions (since an additional 118 samples are classified correctly by the robust model and now fall in the red and blue distributions). The macro component seeks a representation that remains invariant to alterations from $\delta$ at the distribution level. We model this with the following $\mathcal{L}_{{Dist}}(H^{B}_{\textbf{X}}, H^{B}_{\textbf{X} + \delta})$.



\begin{equation}
    \label{equ:align_loss}
    \begin{split}
        \min_{\theta}\mathbb{E}_{\{x,y\}\sim D}\Bigg[\underset{\text{for accuracy}}{ \underbrace{ \mathcal{L}(f_{\theta}(\textbf{X},y))}} \\ + 
        \max_{ d(g(\textbf{X} ),g(\textbf{X+$\delta$})) \le \epsilon} \underset{\text{for robustness}}{\underbrace{\mathcal{L}_{{Dist}}(H^{B}_{\textbf{X}}, H^{B}_{\textbf{X} + \delta})}} \Bigg]  
    \end{split}
\end{equation}
 
In equation \ref{equ:align_loss}, $H^{A}_{\hat{\textbf{X}}} = H^{B}_{\textbf{X} + \delta}$ represents the distribution from the base domain that has been perturbed with $\delta$, and $H^{B}_{\textbf{X}}$ denotes the distribution from the base domain itself.

\subsection{Aligned semantic robust NLP model}

Our first objective is the further finetuned model on the target dataset, that anchors the performance of the model when no adversarial objective is present. It is defined by a loss function when doing inference with $x$ and $y$. In our experiments, we use the cross entropy loss. The second (Macro) objective formulate the classical saddle point problem commonly found in adversarial training. The second objective employs a distance computation between the representations of the model undergoing base inference and the representations of the mode undergoing adversarial inference.

\begin{equation}
    \label{equ:macro_equation}
    \mathcal{L} = \mathcal{L}_{Base} +  \lambda \mathcal{L}_{Dist}
\end{equation}

Where $\lambda$ controls the strength associated with the distance regularizer. 

\subsection{Approximating $\mathcal{L}_{Dist}$ (Macro) }


There are various methods available for modeling $\mathcal{L}_{Dist}$. The use of a parametric distance measure, such as the Kullback-Leibler (KL) divergence, would be preferable under ideal circumstances. However, implementing this measure in practice is complex because it requires intermediate density estimations within a high-dimensional space, for example, a 768-dimensional embedding as in BERT. Additionally, the complexity is compounded by the small number of samples typically included in each batch, like 64. Non-parametric alternatives encompass methods such as MMD with a RBF kernel \cite{MMD, MMD_Distance}, CORAL \cite{CORAL}, and, akin to the approach described in \cite{OT_Images}, utilizing an optimal transport regularizer applied to output features. An extensive explanation of each distance measure is given in Appendix \ref{distance_measures}.

\section{Experimental Setup}
\subsection{Backbone models and tasks}
We evaluated the \textit{BERT} \cite{BERT} and \textit{RoBERTa} \cite{RoBERTa} models on the classification tasks \textit{MR}, \textit{AG-News}, \textit{SST-2}.
 
\subsection{Evaluation metrics}
We utilize the evaluation framework previously proposed in \cite{TextAttack}, where an evaluation set is perturbed, and we record the following data from the Total Attacked Samples ($TAS$) set: Number of Successful Attacks ($N_{succ-atk}$), Number of Failed Attacks ($N_{fail-atk}$), and Number of Skipped Attacks ($N_{skp-atk}$). We utilize these values to record the following metrics. \textit{Clean accuracy/Base accuracy/Original accuracy}, which offers a measure of the model's performance during normal inference. \textit{After attack accuracy/Accuracy under attack} ($A_{aft-atk}=\frac{N_{fail-atk}}{TAS}$) or (AUA), is critical, representing how effectively the attacker deceives the model across the dataset. Similarly, the \textit{After success rate} ($A_{succ-rte}=\frac{N_{succ-atk}}{TAS-N_{skp-atk}}$) or (ASR) excludes previously misclassified samples. The paper also considers the \textit{Percentage of perturbed words}, the ratio of disturbed words to the total in a sample, and \textit{Semantic similarity}, an automatic similarity index, modelled by $d$ \cite{Universalsentenceencoder}. Lastly, the \textit{Queries} denotes the model's number of invocations for inference. Appendix \ref{evaluation_metrics_extended} and \ref{evaluation_strategy} give more details on our evaluation metrics and strategy.

\subsection{Evaluation baselines}
  The assessment involves comparing to seven baseline techniques. These baselines include: The `Vanilla' model, which is trained solely on the base dataset. `TextFooler + Adv Aug' which incorporates adversarial samples generated by TextFooler into the base dataset as a form of data augmentation. `TextFooler + Adv Reg' which also uses adversarial samples from TextFooler, but instead applies a regularization approach with both $\lambda_{0}$ and $\lambda_{1}$ set to 1, as detailed in "Standard" under Table \ref{defence_regularizers} in the Appendix. Additionally, there are four other baseline models to compare against. Among them is `Attack-to-Train' (A2T), which, according to the latest available information, is currently the leading state-of-the-art (SOTA) method for adversarial training at the token level in the NLP domain, as referenced in \cite{A2T}. Next, there are embedding level adversarial training baselines. Among these there is `InfoBERT', which is recognized for enhancing model robustness by requiring fewer projected gradient steps compared to FreeLB. Following this, we have `FreeLB++', an advancement over FreeLB that achieves performance gains by increasing the $\epsilon$-bound. Finally, there is `DSRM', a novel approach that introduces perturbations whenever the loss dips below a predefined threshold. According to our latest knowledge, FreeLB++ surpasses most existing baselines \cite{TextDefender}. Meanwhile, DSRM, as a more recent innovation, demonstrates superior performance over FreeLB++.


\section{Experiments}

We initially investigate the performance of the system using all the regularizers outlined in Equation \ref{equ:macro_equation} (referred to as Macro). When generating the adversarial candidates using TextFooler, PWWS, BERTAttack or TextBugger we set a angular similarity $\epsilon$, used to assess the semantics of a transformation, to 0.5 where $d$ is the Universal Sentence Transformer \cite{Universalsentenceencoder}, and the number of word substitution candidates $N$ to 50, which are both the standard used in previous work, we also utilize a greedy search approach with word replacement inspired by \cite{Textfooler} to identify suitable word substitutions in a query efficient way.

For the macro component, we undertake a comparative analysis involving adversarial training and attacks utilizing word substitutions derived from various word embeddings (WordNet, Contextual, Counter Fitted, and GloVe). Furthermore, we conduct an in-depth examination of the resultant embeddings through t-SNE to gain valuable insights into the behavior of the feature space.

\section{Results}

\subsection{Distribution alignment through MMD}

In this section, we investigate the effectiveness of a training method that utilizes adversarial data sampled from four different attacks which use different word embeddings: PWWS (WordNet), TextFooler (Counter-Fitted), BERTAttack (Contextual) and TextBugger (GloVe). Our primary objective is to test whether a distance regularizer limits the word level perturbations from creating samples residing in the adversarial distributions shown in Figure \ref{img:aug_data_distribution} (top) with color yellow and cyan and secondly to evaluate if the new robust algorithm generalizes across word embeddings since the underlying hypothesis is that by leveraging a distance metric, which enables successful learning of domain-invariant features, we can anticipate improved robustness not only against the same attack but also across different attacks. This assumption is grounded in the fundamental similarities shared by word-level attacks. 
Additionally, Table \ref{distribution_alignment_mmd} provides a comprehensive overview of the performance gains observed in terms of after-attack accuracy (AUA) and the corresponding drop in the attack success rate (ASR).

\begin{table*}[thp]
\centering
\caption{Adversarial training using distribution alignment on MR. Values in \textbf{Bold} represent the highest scores, those in round brackets ($*$) denote the second highest, and values in square brackets [$*$] indicate the third highest. Extended results are in Table \ref{Appendix:distribution_alignment_mmd} in the Appendix.}
\label{distribution_alignment_mmd}
\scalebox{0.5}{
\begin{tabular}{lllcccccccccccc}
\hline
\multicolumn{2}{l}{\multirow{3}{*}{\textbf{Model}}} & \multirow{3}{*}{\textbf{\begin{tabular}[c]{@{}l@{}}Train method \\ (Defense)\end{tabular}}} & \multicolumn{12}{c}{\textbf{Test Method}} \\
\multicolumn{2}{l}{} &  & \multicolumn{3}{c}{\textbf{\begin{tabular}[c]{@{}c@{}}PWWS  \\ (WordNet)\end{tabular}}} & \multicolumn{3}{c}{\textbf{\begin{tabular}[c]{@{}c@{}}BERTAttack  \\ (Contextual)\end{tabular}}} & \multicolumn{3}{c}{\textbf{\begin{tabular}[c]{@{}c@{}}TextFooler \\ (Counter Fitted)\end{tabular}}} & \multicolumn{3}{c}{\textbf{\begin{tabular}[c]{@{}c@{}}TextBugger\\ (Sub-W GloVe)\end{tabular}}} \\
\multicolumn{2}{l}{} &  & \textbf{CA (↑)} & \textbf{AUA (↑)} & \textbf{ASR (↓)} & \textbf{CA (↑)} & \textbf{AUA (↑)} & \textbf{ASR (↓)} & \textbf{CA (↑)} & \textbf{AUA (↑)} & \textbf{ASR (↓)} & \textbf{CA (↑)} & \textbf{AUA (↑)} & \textbf{ASR (↓)} \\ \hline
\multicolumn{2}{l}{\multirow{11}{*}{\textbf{BERT}}} & Vanilla & 86.0 & 21.2 & 75.35 & 86.0 & 32.6 & 62.09 & 86.0 & 13.8 & 83.95 & 86.0 & 5.8 & 93.26 \\
\multicolumn{2}{l}{} & TextFooler + Adv Aug & 85.6 & 21.0 & 75.47 & 85.6 & 31.6 & 63.08 & 85.6 & 11.8 & 86.21 & 85.6 & 4.0 & 95.33 \\
\multicolumn{2}{l}{} & TextFooler + Adv Reg & 86.8 & 25.8 & 70.28 & 86.8 & 34.2 & 60.6 & 86.8 & 18.4 & 78.8 & 86.4 & 8.0 & 90.74 \\
\multicolumn{2}{l}{} & A2T & 86.8 & 18.0 & 79.26 & 86.8 & 32.6 & 62.44 & 86.8 & 13.8 & 84.1 & 86.8 & 3.8 & 95.62 \\
\multicolumn{2}{l}{} & InfoBERT & 82.6 & 31.4 & 61.99 & 82.6 & 34.6 & 58.11 & 82.6 & 19.6 & 76.27 & 82.6 & 6.8 & 91.77 \\
\multicolumn{2}{l}{} & FreeLB++ & 84.2 & 26.6 & 68.41 & 84.2 & 32.4 & 61.52 & 84.2 & 16.4 & 80.52 & 84.2 & 3.8 & 95.49 \\
\multicolumn{2}{l}{} & DSRM & 87.6 & 25.6 & 70.78 & 87.6 & 32.6 & 62.79 & 87.6 & 20.0 & 77.17 & 87.6 & 11.0 & 87.44 \\ \cline{3-15} 
\multicolumn{2}{l}{} & PWWS + MMD (SemRoDe) & 86.0 & 37.0 & 56.98 & 86.0 & {[}45.8{]} & 46.74 & 86.0 & 38.2 & 55.58 & 86.0 & 28.2 & 67.21 \\
\multicolumn{2}{l}{} & BERTAttack + MMD (SemRoDe) & 85.8 & \textbf{38.2} & 55.48 & 85.8 & (46.6) & 45.69 & 85.8 & \textbf{40.0} & 53.38 & 85.8 & \textbf{29.4} & 65.73 \\
\multicolumn{2}{l}{} & TextBugger + MMD (SemRoDe) & 85.8 & {[}37.4{]} & 56.41 & 85.8 & \textbf{47.6} & 44.52 & 85.8 & (39.4) & 54.08 & 85.8 & (28.8) & 66.43 \\
\multicolumn{2}{l}{} & TextFooler + MMD (SemRoDe) & 85.8 & (37.8) & 55.94 & 85.8 & 44.6 & 48.02 & 85.8 & {[}39.0{]} & 54.55 & 85.8 & {[}28.6{]} & 66.67 \\ \hline
\multicolumn{2}{l}{\multirow{11}{*}{\textbf{RoBERTa}}} & Vanilla & 89.6 & 24.0 & 73.21 & 89.6 & 33.2 & 62.95 & 89.6 & 12.0 & 86.61 & 89.6 & 5.0 & 94.42 \\
\multicolumn{2}{l}{} & TextFooler + Adv Aug & 90.0 & 30.4 & 66.22 & 90.0 & 36.0 & 60.0 & 90.0 & 21.4 & 76.22 & 90.0 & 9.0 & 90.0 \\
\multicolumn{2}{l}{} & TextFooler + Adv Reg & 90.8 & 31.8 & 64.98 & 90.8 & 34.6 & 61.89 & 90.8 & 21.0 & 76.87 & 90.8 & 9.4 & 89.65 \\
\multicolumn{2}{l}{} & A2T & 92.0 & 24.2 & 73.7 & 92.0 & 28.8 & 68.7 & 92.0 & 15.0 & 83.7 & 92.0 & 4.4 & 95.22 \\
\multicolumn{2}{l}{} & InfoBERT & 89.2 & 30.4 & 65.92 & 89.2 & 29.4 & 67.04 & 89.2 & 19.4 & 78.25 & 89.2 & 4.4 & 95.07 \\
\multicolumn{2}{l}{} & FreeLB++ & 91.6 & 29.4 & 67.9 & 91.6 & 33.8 & 63.1 & 91.6 & 20.0 & 78.17 & 91.6 & 6.6 & 92.79 \\
\multicolumn{2}{l}{} & DSRM & 88.4 & 29.0 & 67.19 & 88.4 & 39.8 & 54.98 & 88.4 & 25.2 & 71.49 & 88.4 & 9.8 & 88.91 \\ \cline{3-15} 
\multicolumn{2}{l}{} & PWWS + MMD (SemRoDe) & 89.4 & {[}53.2{]} & 40.49 & 89.4 & \textbf{57.8} & 35.35 & 89.4 & (55.4) & 38.03 & 89.4 & (43.2) & 51.68 \\
\multicolumn{2}{l}{} & BERTAttack + MMD (SemRoDe) & 89.8 & \textbf{53.6} & 40.31 & 89.8 & {[}56.8{]} & 36.75 & 89.8 & 46.2 & 48.55 & 89.8 & \textbf{56.4} & 37.19 \\
\multicolumn{2}{l}{} & TextBugger + MMD (SemRoDe) & 89.4 & (53.4) & 40.27 & 89.4 & (57.2) & 36.02 & 89.4 & \textbf{55.6} & 37.81 & 89.4 & {[}42.8{]} & 52.13 \\
\multicolumn{2}{l}{} & TextFooler + MMD (SemRoDe) & 88.8 & 51.8 & 41.67 & 88.8 & 56.6 & 36.26 & 88.8 & {[}54.8{]} & 38.29 & 88.8 & 42.2 & 52.48 \\ \hline
\end{tabular}
}
\end{table*}

The insights derived from the presented analysis are twofold. Firstly, as depicted in Table \ref{distribution_alignment_mmd}, 
Within this context, the indispensability of distribution alignment becomes evident, as it consistently leads to significant performance gains in after-attack accuracy, regardless of the employed word embedding. Secondly, the observed improvements in performance remain consistent even when subjected to various word embedding attacks. This robustness across diverse attacks further affirms the efficacy and generalizability of the achieved performance increments.

\subsection{Effect of distribution alignment method}

We perform an ablation study on the choice of distance metric used for distribution alignment Table \ref{vs_all_plot_over_epochs} showcases MMD, Coral, and Optimal Transport. We found training on optimal transport to be unstable and only effective for a narrow set of hyperparameters. For other attackers, MMD works more reliably and has better after-attack performance, achieving AUA of 37.8\% ,44.6\%, 39\%, and 28.66\% for PWWS, BertAttack, TextFooler, and TextBugger, respectively. Although Coral and Optimal Transport don't perform as well as MMD, they still outperform the baselines when tested on the MR dataset. Because of these results, we focus our experimentation with MMD. This might be due to MMD being a method that aligns lower statistical components, compared to CORAL and Optimal Transport, which may help when training using batches, in fact, it was shown that optimal transport does indeed suffer when the transport map isn't computed across all samples in a dataset \cite{Optimal_Transport_Limitation}.

\begin{table}[h]
\centering
\caption{Model (BERT) performances for AUA depending on the alignment regularizer when trained on the MR dataset. We use TextFooler to generate the adversarial samples that form the adversarial domain.}
\label{vs_all_plot_over_epochs}
\scalebox{0.43}{
\begin{tabular}{lll|cccccccccccc}
\hline
\multicolumn{3}{l}{\multirow{3}{*}{\textbf{\begin{tabular}[c]{@{}l@{}}Distance\\ Metric\end{tabular}}}} & \multicolumn{12}{c}{\textbf{Test Method (MR/BERT) Where CA (↑) AUA (↑) ASR (↓)}} \\
\multicolumn{3}{l}{} & \multicolumn{3}{c}{\begin{tabular}[c]{@{}c@{}}PWWS \\ (WordNet)\end{tabular}} & \multicolumn{3}{c}{\begin{tabular}[c]{@{}c@{}}BERTAttack  \\ (Contextual)\end{tabular}} & \multicolumn{3}{c}{\begin{tabular}[c]{@{}c@{}}TextFooler \\ (Counter Fitted)\end{tabular}} & \multicolumn{3}{c}{\begin{tabular}[c]{@{}c@{}}TextBugger\\ (Sub-W GloVe)\end{tabular}} \\ \cline{4-15} 
\multicolumn{3}{l}{} & CA & AUA & ASR & CA & AUA & ASR & CA & AUA & ASR & CA & AUA & ASR \\ \hline
\multicolumn{3}{l}{Vanilla} & 86.0 & 21.2 & 75.35 & 86.0 & 32.6 & 62.09 & 86.0 & 13.8 & 83.95 & 86.0 & 5.8 & 93.26 \\
\multicolumn{3}{l}{L2 Distance} & 85.4 & 14.6 & 82.9 & 85.4 & 26.6 & 68.85 & 85.4 & 8.0 & 90.63 & 85.4 & 2.4 & 97.19 \\
\multicolumn{3}{l}{CORAL (SemRoDe)} & 86.0 & 27.0 & 68.6 & 86.0 & 37.8 & 56.05 & 86.0 & 21.0 & 75.58 & 86.0 & 10.2 & 88.14 \\
\multicolumn{3}{l}{OT (SemRoDe)} & 85.6 & 37.4 & 56.31 & 85.6 & 35.6 & 58.41 & 85.6 & 28.6 & 66.59 & 85.6 & 13.4 & 84.35 \\
\multicolumn{3}{l}{MMD (SemRoDe)} & 85.8 & \textbf{37.8} & 55.94 & 85.8 & \textbf{44.6} & 48.02 & 85.8 & \textbf{39.0} & 56.97 & 85.8 & \textbf{28.6} & 66.67 \\ \hline
\end{tabular}
}

\end{table}

\subsection{Why distribution alignment works}
\begin{figure}[th]
    \centering
    \includegraphics[width=.78\linewidth]{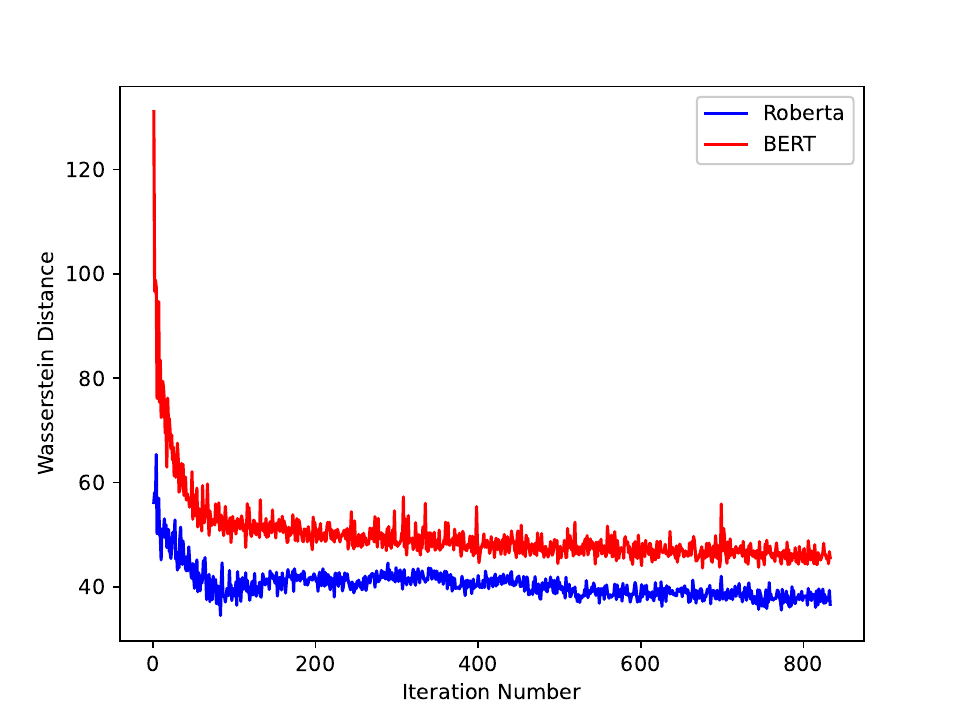} 
    \includegraphics[width=.78\linewidth]{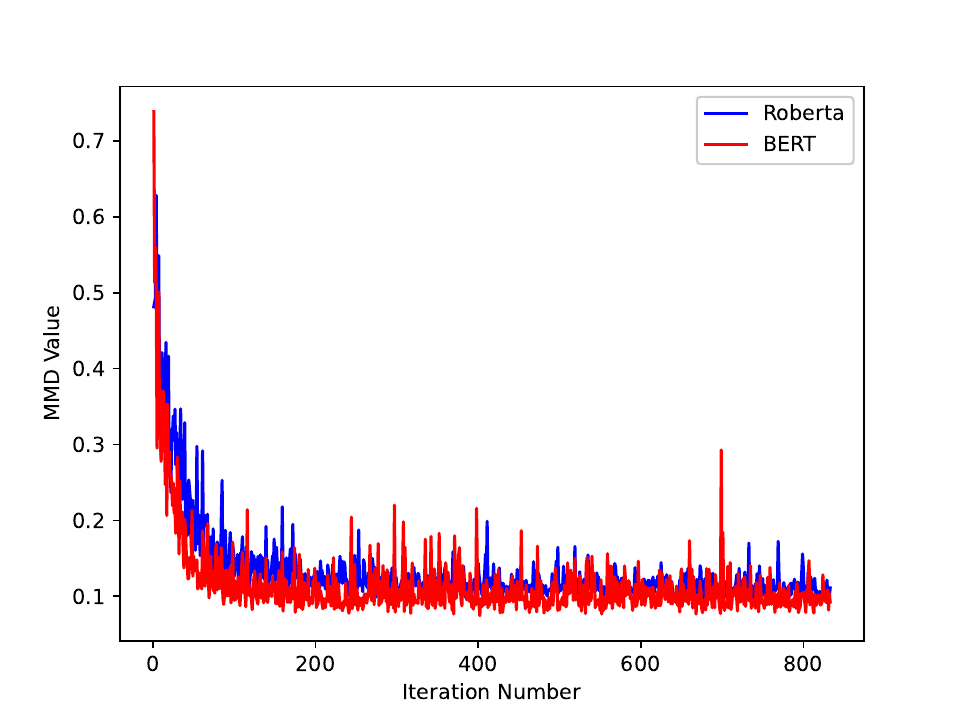} 
    \caption{ OT (Top) MMD (Bottom) response over iterations for MR
    }
    \label{img:distance_comparison_plot}
\end{figure}

Table \ref{distribution_alignment_mmd} illustrates how the after-attack accuracy of a model, trained on word substitutions generated from different embedding sets such as counter-fitted embeddings (TextFooler) \cite{Counter_Fitted_Embeddings}, WordNet (PWWS) \cite{Word_Net}, or masked language modeling (BERTAttack), can be effectively generalized across these sets. This observation can be attributed to the presence of a base domain (represented by the red/blue cluster) and an adversarial domain (represented by the yellow/cyan cluster) as depicted in Figure \ref{img:aug_data_distribution}. Consequently, training the model with an objective that aligns these two distributions and creates a new representation in the feature space can result in improved robustness performance. This can be ascertained by the Wasserstein distance dropping over time Figure \ref{img:distance_comparison_plot} (top), when training with Equation \ref{equ:macro_equation}, this drop, which represents the two hypothetical base and adversarial feature distributions becoming similar, can be seen despite utilizing MMD as a distance measure. Naturally, this occurs in a high dimension, which makes it difficult to show using a dimensionality reduction technique such as t-SNE in Figure \ref{img:aug_data_distribution}. Interestingly, in Figure \ref{img:distance_comparison_plot} the Wasserstein distance seem to capture the relationship between distributions in a smoother way than using MMD, this is further shown in other datasets such as AGNEWS and SST2 in Appendix \ref{Appendix:distribution_alignment_mmd}.

\subsection{Interpreting Distribution Alignment }
To provide insights into the path an example takes in the feature space as it becomes adversarial, we employ t-SNE visualization. This approach allows us to track the trajectory of samples as they undergo perturbations, which has remained largely unexplored in the existing literature. With Figure \ref{img:aug_data_distribution} and Figure  \ref{img:TSNE_OT_Non_Robust74}, \ref{img:TSNE_OT_Robust74} in the Appendix we present the results of our experiments. Figure \ref{img:aug_data_distribution} (top) provides insight into the distributions within the base model. Moving forward, we explore the model that incorporates a distance metric (MMD) as a regularizer, depicted in Figure \ref{img:aug_data_distribution} (bottom). This picture present the distribution components of the model utilizing MMD. These visualizations collectively contribute to our understanding of the behavior and transformations occurring within the feature space. There are three important extrapolations to consider from these figures. Firstly, the robust model exhibits fewer adversarial samples (cyan and yellow points) due to a higher number of correctly classified points. Secondly, by quantifying the number of perturbations required to successfully perturb a sample across the boundary (illustrated by the black samples with arrows), we observe most samples, to successfully be perturbed across a classification boundary require more word substitutions. The black point shows this. Originally in Figure \ref{img:aug_data_distribution} (top) only perturbing one word was sufficient to achieve a successful attack. In Figure \ref{img:aug_data_distribution} (bottom) the same sample needs to have 4 word substitutions before being successfully perturbed (Positive to Negative sentiment in Table \ref{Appendix:distribution_alignment_mmd} in the appendix).

\subsubsection{Qualitative samples}
After a sufficient number of iterations ($\delta$ steps) on an embedding, FreeLB++ causes the embedding to move closer to a neighboring discrete embedding. This, in turn, leads to a change in token when mapping the second embedding, as demonstrated in Table \ref{Qualitative_sample}. While the token substitution has the potential to preserve synonym information, there is no guarantee. We observed that training with too many $\delta$ steps, resulting in frequent token substitutions, actually worsens the performance of adversarial training. This finding is consistent with the observations made in \cite{TextDefender}, emphasizing a significant drawback of FreeLB++. We believe this issue can be addressed by sampling adversarial samples using high-quality word embeddings and by matching the base and adversarial distributions.

\begin{table}
    \centering
    \scalebox{0.6}{
    \begin{tabular}{c c c c c c c c c c c c c c}
        \multicolumn{1}{c}{\textbf{Type}} & \multicolumn{11}{c}{\textbf{Sample}} \\ \hline
        Original & \multicolumn{11}{l}{\begin{tabular}[c]{@{}l@{}}\textbf{\color{blue}{{[}CLS{]}}} Woods' Top Ranking on Line at NEC Invite \\ (AP) AP - Tiger Woods already lost out on the majors. \\Next up could be his No. 1 ranking. \textbf{ \color{blue}{{[}SEP{]}}}\end{tabular}} \\ \hline
        FreeLB++ & \multicolumn{11}{l}{\begin{tabular}[c]{@{}l@{}}\textbf{ \color{blue}{ {[}CLS{]}}} Woods' Top Ranking on Line at NEC Invite \\ \textbf{\color{red}{{[}SEP{]}}} (AP) AP - Tiger Woods \textbf{\color{red}{strong}} lost \textbf{\color{red}{partners love}} \\ the majors \textbf{\color{red}{and}} \textbf{\color{red}{{[}SEP{]}}} up \textbf{\color{red}{xiao ko .}} No. 1 ranking \textbf{\color{red}{to won}}\end{tabular}} \\ \hline
        TextFooler & \multicolumn{11}{l}{\begin{tabular}[c]{@{}l@{}}\textbf{\color{blue}{{[}CLS{]}}} Woods' Keynote \textbf{\color{green}{Rank}} on \textbf{\color{green}{Routing}} at NEC Invite \\ (AP) AP - Tiger Woods already lost out on the \\ \textbf{\color{red}{commandant}}. Next up could be his No. 1 ratings. \textbf{\color{blue}{{[}SEP{]}}}\end{tabular}} \\ \hline
    \end{tabular}
    }
    \caption{Example of a qualitative sample generated by FreeLB after a large amount of $\delta$ steps. We use an initial $\delta$ of size 0.05 with a gradient ascend step size of 5 for about 30 steps. \color{red}{Red: Bad change, embedding method}\color{black}{,} \color{green}{Green: Good change, conceptual method}\color{black}{.}}
    \label{Qualitative_sample}
\end{table}

\section{Computation time}
SemRoDe is computationally efficient. In Table \ref{table:computation_cost}, the `Adversarial Set Generation' column tracks the time taken to adversarially generate 10\% of the data, while the `Train' column records the time required to train the model. Techniques for embedding perturbation such as FreeLB++, InfoBERT, and DSRM are more time-consuming due to, at times, undergoing multiple PGD steps. In contrast, token perturbation techniques like TextFooler and A2T do not undergo this operation. Table \ref{table:computation_cost} provides a comparative analysis of the computational times for the various methods. Further discussion on computational cost is presented in
Appendix \ref{appendix:computation_time}.

\begin{table}[h]
\centering
\caption{Computation time to generate adversarial samples and perform training on MR with 7 epochs on BERT. Robustness values are shown in Table \ref{distribution_alignment_mmd}}
\label{table:computation_cost}
\scalebox{0.65}{
\begin{tabular}{lcc}
\\ \hline
\multirow{2}{*}{\textbf{Type}} & \multicolumn{2}{c}{\textbf{GPU Time}}   \\ \cline{2-3}
 & \multicolumn{1}{c}{\textbf{\begin{tabular}[c]{@{}c@{}}Adversarial Set\\ Generation\end{tabular}}} & \multicolumn{1}{c}{\textbf{Train}}    \\ \hline

Baseline & - & 365 \\
TextFooler + Aug & 494 & 396 \\
TextFooler + AT & 473 & 691 \\
TextFooler + MMD (SemRoDe) & 537 & 694 \\
A2T & 2655 & 670 \\
DSRM & - & 883 \\
FreeLB++ & - & 4648 \\
InfoBERT & - & 5404 \\
\hline
\end{tabular}
}

\end{table}

\section{Discussion}
Recent studies have taken a novel approach by investigating the impacts of adversarial attacks and defenses on smaller encoding and decoding models, such as BERT, RoBERTa, and GPT. Given that these models are less computationally demanding, they serve as more tractable proxies for larger, cutting-edge models. Promisingly, attacks that work on these models have been shown to be applicable to sophisticated LLMs, according to recent works \cite{Adversarial_LLM3, Adversarial_LLM2, Adversarial_LLM1}. Our research provides insight into aligning the base and adversarial domains and how doing so can improve robust accuracy. As the landscape of LLMs continues to evolve and expand, we hope these findings can be utilized and applied to larger, more sophisticated language models to improve their robustness to adversarial attacks.

\section{Conclusion}
Drawing inspiration from previous work on image settings utilizing continuous data, we demonstrate the existence of a base and adversarial domain within a linguistic setting that uses discrete data. This becomes apparent when an attacker manipulates a system through word-level substitutions. Striving to decrease the divergence between these two domains in the belief that it would improve robust generalization, we employ a distance regularizer on high-level features. This allows for learning an internal representation that restricts attackers from shifting a base sample into the adversarial domain. This method introduces a novel distance-based regularizer, $\mathcal{L}_{Dist}$, as we investigate different distance measures for aligning the domains. Aligning these domains results in a reduced Wasserstein distance and smoother clustering of the t-SNE output features. This procedure, aligning base and adversarial features, fosters robust generalization, as the learned representations maintain robust accuracy despite attackers leveraging word substitutions from four various word-embeddings.

\clearpage

\section{Acknowledgements}
This research/project is supported by the National Research Foundation, Singapore under its Industry Alignment Fund – Pre-positioning (IAF-PP) Funding Initiative. Any opinions, findings and conclusions or recommendations expressed in this material are those of the author(s) and do not reflect the views of National Research Foundation, Singapore.

\section{Limitations}
We do not consider the class-conditional case, and the domains are `inverted'. This means a distance measure like MMD won't necessarily align the adversarial samples of class one with its respective samples from the base class. Instead, it will simply reduce the distance between the base and adversarial domains, which might impede robust performance. Therefore, in future work, it will be necessary to conduct a study where the objective takes classes into account when performing $\mathcal{L}_{Dist}$. Furthermore, we currently do not perform online adversarial training, where we generate adversarial samples in every batch. This could be suboptimal as the model is dynamic and changes with each batch. Another limitation involves robustness against white-box, embedding perturbations. Considering our threat model, we do not anticipate that attackers will have access to the embedding space. Therefore, the model will lack robustness against any adversarial perturbation that results in an input embedding not directly mappable back to the input space. This is because the adversarial training algorithm will not have been exposed to such data points. Lastly, how this alignment technique can be extended to generative models remains unexplored.

\section{Ethics Statement}
This research was conducted in accordance with
the ACM Code of Ethics.

\bibliography{anthology,custom}
\bibliographystyle{acl_natbib}

\newpage
\appendix

\section{Implementation details}\label{Implementation_details}
In the case of the SemRoDe macro model, adversarial examples are generated during the first epoch. We generate adversarial samples where each word in a sample has $N$ substitution candidates, for each candidate substitution the greedy search with word replacement algorithm checks whether such substitution reduces the classification confidence at the logit, if the transformation is successful, it needs to achieve above $\epsilon=0.5$ angular similarity to be considered a successful adversarial sample, which will be included in training. The difference between the TextFooler, BERTAttack, PWWS and TextBugger methods is the utilized word embeddings to generate $N$. The MR model is trained on this data for seven epochs, while both the AGNews and SST2 models are trained for three epochs each.

\subsection{Offline vs online adversarial training}
Generating the dataset one time at the start, which we call offline adversarial training, differs from typical adversarial training techniques, which create adversarial samples at every epoch or batch to consider the model's dynamic learning nature. We have found this method best given that the generation of adversarial samples through TextFooler, BERTAttack, PWWS, and TextBugger can be time-consuming due to the heuristic nature of these attacks.

Naturally, we experiment with online adversarial training on BERT/MR, following the conventional approach of dynamically generating adversarial samples at each epoch. However, we observed no significant improvement in performance. Our conclusion is that epoch-wise adversarial training may not be necessary, and that an adversarial data augmentation strategy that follows an offline adversarial training paradigm is sufficient to align base and adversarial representations, thereby ensuring robustness against word-level attacks. Additionally, this approach offers the benefit of being a more rapid training setting, as detailed in Table \ref{table:computation_cost}.

\subsection{Baseline implementation}
Nonetheless, we continue to reference state of the art baselines in our comparative analysis. The FreeLB++ model remains unmodified as it generates adversarial samples at the embedding level at each batch for all data. As for A2T, we preserve its original implementation. The entire training data is exposed to the algorithm, which then generates approximately 20\% of the data as adversarial samples. This percentage aligns with the recommendations outlined in the original implementation. 

We implemented InfoBert in the same manner as it was in TextDefender; however, it should be noted that with the original learning rate of 2e-5 for the inner steps, in our case, the algorithm was unable to converge to a robust model. Consequently, we experimented with a learning rate of 0.1, which represents the upper limit recommended by the original implementation of InfoBert \cite{InfoBERT}. Additionally, we increased the number of projected gradient descent (PGD) steps from 3 to 7 for both AGNEWS and SST2 datasets across all models. For the MR dataset, which is smaller, we conducted 15 steps on BERT and 25 on ROBERTA. With the exception of these modifications, all other hyperparameters remained unchanged.

When it came to implementing the DSRM \cite{DSRM}, we adhered closely to the original publication, adjusting only the clipping parameter. The original clipping value prescribed in the paper and code, in our experiments, did not yield an improved after attack accuracy; thus, we set the loss clamp to 50 for both SST2 and AGNEWS. In practice, this adjustment is akin to eliminating the clamping entirely. As a result, the algorithm modifies the batch at every iteration. We observed that this approach is able to enhanced the accuracy after the attack while also largely maintaining the original accuracy. For the MR dataset, however, such a high loss clamp rendered the algorithm ineffective. Therefore, we adjusted the clamp to 0.5. As we trained BERT and ROBERTA models from scratch utilizing DSRM, we extended the training to 10 epochs for AGNEWS and SST2, and 50 epochs for MR, choosing the model that demonstrated the best performance.

\subsection{Datasets}
We Test on 3 classification datasets, the base distribution size and adversarial distribution size are presented in Table \ref{Dataset_sizes}.

\begin{table}[ht]

\centering
\caption{Dataset sizes}
\label{Dataset_sizes}
\scalebox{0.70}{
\begin{tabular}{llcccc}
\hline
\textbf{Task} & \textbf{Dataset} & \textbf{Train} &  \begin{tabular}[c]{@{}c@{}}Adv Data \\ Size (Train)\end{tabular}  & \textbf{Test} & \textbf{Classes} \\ \hline
\multirow{2}{*}{\begin{tabular}[c]{@{}l@{}}Sentiment \\ Classification\end{tabular}} & MR & 8.5k & 850 & 1k & 2 \\ \cline{2-6}
 & SST-2 & 67k & 6.7k & 1.8k & 2 \\ \hline
\begin{tabular}[c]{@{}l@{}}News \\ Classification\end{tabular} & AGNEWS & 120k & 12k & 7.6k & 4 \\ \hline
\end{tabular}}
\end{table}



\section{Evaluation metrics extended}\label{evaluation_metrics_extended}
We utilize the evaluation framework previously proposed in \cite{TextAttack}, where an evaluation set is perturbed, and we record the following metrics from the Total Attacked Samples ($TAS$) set: Number of Successful Attacks ($N_{succ-atk}$), Number of Failed Attacks ($N_{fail-atk}$), and Number of Skipped Attacks ($N_{skp-atk}$). The \textit{Clean accuracy/Base accuracy/Original accuracy} represents the accuracy of the model undergoing normal inference. A high clean accuracy indicates that the model performs well for its intended task. The \textit{After attack accuracy/Accuracy under attack} ($A_{aft-atk}=\frac{N_{fail-atk}}{TAS}$) or (\textbf{AUA}) is the most crucial metric, representing how effectively the attacker can deceive the model across the dataset. Lower values of AUA indicate a higher success rate in fooling the model. The \textit{After success rate} ($A_{succ-rte}=\frac{N_{succ-atk}}{TAS-N_{skp-atk}}$) or (\textbf{ASR}) is similar to AUA but excludes previously misclassified samples. The \textit{Percentage of perturbed words} refers to the ratio of perturbed words to the total number of words in the sample. This metric should be minimized as perturbing more words makes the sample's manipulation more detectable. \textit{Semantic similarity} \cite{Textfooler,HardLabel} is an automatic similarity index that quantifies the visual difference between two samples using a deep learning model. In this case, modelled by $d$, the Universal Sentence Encoder \cite{Universalsentenceencoder} is utilized along with a angular cosine similarity hyper parameter $\epsilon$ between the output embeddings. An output value of 1 indicates semantic equivalence, while 0 represents no similarity. \textit{Queries} denotes the number of times the algorithm needs to invoke the model for inference; keeping this low helps avoid detection. We further explain the evaluation strategy in Appendix \ref{evaluation_strategy}.

\section{Evaluation Strategy}\label{evaluation_strategy}
The experiments section will follow the evaluation setting originally proposed by TextDefender \cite{TextDefender}. The main difference is that we will still test whether a method can generalize when using the same synonym set. This testing is relevant due to the requirement of learning robust models against existing attacks, such as BERTAttack and PWWS. Additionally, we propose another evaluation test that assesses the generalization of training on one set of synonyms against other sets. This test is applicable because these attacks already exist, but we do not yet have training strategies to address them. It will also demonstrate how well training on one set of synonyms can generalize to attackers using other synonym sets. In line with the technique adopted by TextDefender, we set the $\epsilon$ value to $0.5$, leading to a semantic similarity threshold of $0.84$. In a move to achieve balance, we ensure that the perturbation does not exceed 30\% of the words in each sample. Concurrently, we adhere to a maximum limit of $L*N$ queries per sample, with $L$ denoting the length of the sample in words. As the parameters \textit{Queries}, \textit{Percentage of perturbed words}, and \textit{Semantic similarity} maintain constancy, we have sidestepped their inclusion in our tables. 

\section{Further explanation of adversarial training in language models}
The technique for creating adversarial examples is crucial and our work differs in this respect from previous work such as Contextualized representation-Adversarial Training (CreAT) \cite{CreAT}. This work applies PGD over 'k' steps with an L2-norm restriction on the perturbation magnitude, which parallels prior research, such as \cite{L2ball1,L2ball2}. However, contemporary word-level attacks do not adhere to the L2-norm constraint. Instead of adding small delta perturbations, these attacks use word substitutions, with semantic similarity serving as the boundary for perturbations. This strategy allows an adversary to more readily bypass the defender's constraints. Consequently, we are faced with the critical question: How can we incorporate these same bounds and deltas into our adversarial training technique? The most practical solution is to enforce these restrictions at the input stage by adopting the approaches utilized in well-established attack mechanisms. Hence, our method is superior as it enables the algorithm to more thoroughly explore the attack space. Compared to CreAT we also are attempting to align distributions, and therefore analyse them in detail. Different from our work, CreAT introduces a technique to better train the encoder part of a model with adversarial training, since they noticed normal adversarial training struggles to do so, especially when applied to a encoder-decoder model. In our work, when incorporating samples generated through embedding perturbation techniques into regular adversarial training, the performance improvement is relatively minor. We hypothesize that this is due to the inherent discrepancy between the base and adversarial samples, which belong to two distinct distributions within the feature space. Standard adversarial training does not adequately address this divergence, thus highlighting the necessity of a distance regularizer to better align these two distributions.

\section{Explanation of $\epsilon$}
In our study, the angular similarity $\epsilon$ is employed as a hyperparameter to place constraints on adversarial samples generated with algorithm $g$ in conjunction with $d$. Model $d$ can be any qualitative variant capable of preserving the fundamental attributes of the base sample such as meaning, fluency, grammar, similarity, and so on. The pivotal area of interest tends to be semantic similarity, often represented by the output from the universal sentence encoder. We establish our threshold by employing the formula $w=1-\frac{\epsilon}{\pi}$, where $\epsilon$ has the bounds (0,$\pi$). In this study, we have chosen $\epsilon = 0.5$ which yields a threshold of $0.84$. Table \ref{table:ablation_epsilon} demonstrates how employing different values of $\epsilon$ can result in varying intensity of attacks. Note that a small $\epsilon$ results in a small angle between the original sample and adversarial sample, hence, if any adversarial sample from $g$, after being processed by $d$ is above this $\epsilon$, the sample is rejected. 

Previous research has primarily maintained a fixed $\epsilon$ in the vicinity of word embeddings, where the structure of the boundary is usually in the form of either an l2-ball or a convex hull. In our experiments, we leverage heuristic adversarial samples bound by $d$ with the aim to restrict the adversaries not based on arbitrary objectives such as an l2 ball, but rather, on important aspects of linguistic characteristics like semantic similarity.

Also take note that increasing N and reducing $\epsilon$ have similar effects. They both lead to the generation of more diverse adversarial samples that can further reduce the classification confidence. However, referring to the Epsilon and Embedding ablation study (Table \ref{table:ablation_epsilon} and Table \ref{table:ablation_embeddings}), this diversity often comes at the cost of quality, because the semantics are liable to significant alterations.

\section{Large $\delta$ Explanation}\label{large_delta_explenation}
\begin{figure}[thp]
    \centering
    \includegraphics[width=.8\linewidth]{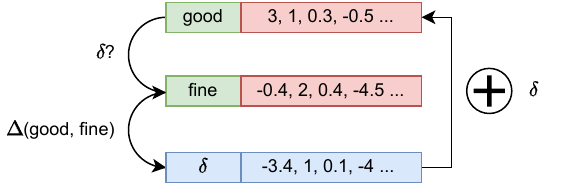} 
    \caption{
        Doing a word substitution is the same as adding a large $\delta$ of fixed size to each word pair. Normally in adversarial training $\delta$ is set to 0.5, this, in comparison, is a small perturbation.
    }
    \label{img:large_delta_explenation}
\end{figure}

In comparison to the image domain or the addition of a static $\delta$ within the word embedding space, word substitutions in the embedding space incorporate a substantially large, word pair-specific $\delta$, which is challenging to both predict and pre-set (as illustrated in Figure \ref{img:large_delta_explenation}). This potentially exacerbates the divergence between the base domain and the adversarial domain could potentially explain the high starting Wasserstein distance. It might also shed light on the considerable improvement in post-attack accuracy following the minimization of $L_{Dist}$.

\section{Vocabulary and Word Substitution Overlap}
In this section, we emphasize the significant overlap in vocabulary and dictionaries among popular word embeddings (see Table \ref{Exploring_Word_Substitution_Overlap_By_Word_Embedding}). However, we observe a limited overlap in potential word substitutions on a per-word basis (see Table \ref{Vocabulary_Dictionary_Overlap_By_Word_Embedding_Types}). Assuming we have two vocabularies $V_1$ and $V_2$ with $W_1 = W_2 = W$ overlapping tokens, each token $W$ has a set of potential token substitutions denoted as $W^{'}$, where $W^{'}_1$ is unlikely equal to $W^{'}_2$ since these token substitutions are sourced by independent word substitution algorithms or data structures. To calculate the word substitutions overlap, we measure the Jaccard index $J$ using the formula: $J=   \frac{|W^{'}_1 \cap W^{'}_2|}{|W^{'}_1 \cup W^{'}_2|}$. We then compute the average result for each word. Table \ref{Vocabulary_Dictionary_Overlap_By_Word_Embedding_Types} also highlights the $J$ for the $W$ that achieved the highest overlap across the vocabulary in brackets.

\begin{table}
\centering
\scalebox{0.50}
{

\begin{tabular}{lllccccccccc}
 \hline
\multicolumn{12}{c}{\textbf{Exploring Word Substitution Overlap By Word Embedding}} \\ \hline
\multicolumn{2}{l}{} & \multicolumn{10}{c}{Attack Method (Word Embedding)} \\ \cline{3-12} 
\multicolumn{2}{l}{\begin{tabular}[c]{@{}l@{}}Attack Method \\ (Word Embedding)\end{tabular}} & \multicolumn{1}{c|}{Counter-Fitted} & \multicolumn{3}{c|}{WordNET} & \multicolumn{3}{c|}{Contextualized} & \multicolumn{3}{c}{Glove} \\ \hline
\multicolumn{2}{l}{Counter-Fitted} & \multicolumn{1}{c|}{100\%} & \multicolumn{3}{c|}{\begin{tabular}[c]{@{}c@{}}1.6\%\\ (23.4\%)\end{tabular}} & \multicolumn{3}{c|}{\begin{tabular}[c]{@{}c@{}}3.6\%\\ (38\%)\end{tabular}} & \multicolumn{3}{c}{\begin{tabular}[c]{@{}c@{}}10.4\%\\ (73\%)\end{tabular}} \\ \hline
\multicolumn{2}{l}{WordNET} & \multicolumn{1}{c|}{\begin{tabular}[c]{@{}c@{}}1.6\%\\ (23.4\%)\end{tabular}} & \multicolumn{3}{c|}{100\%} & \multicolumn{3}{c|}{\begin{tabular}[c]{@{}c@{}}2\%\\ (11.3\%)\end{tabular}} & \multicolumn{3}{c}{\begin{tabular}[c]{@{}c@{}}2.4\%\\ (22\%)\end{tabular}} \\ \hline
\multicolumn{2}{l}{Contextualized} & \multicolumn{1}{c|}{\begin{tabular}[c]{@{}c@{}}3.6\%\\ (38\%)\end{tabular}} & \multicolumn{3}{c|}{\begin{tabular}[c]{@{}c@{}}2\%\\ (11.3\%)\end{tabular}} & \multicolumn{3}{c|}{100\%} & \multicolumn{3}{c}{\begin{tabular}[c]{@{}c@{}}9\%\\ (81.8\%)\end{tabular}} \\ \hline
\multicolumn{2}{l}{Glove} & \multicolumn{1}{c|}{\begin{tabular}[c]{@{}c@{}}10.4\%\\ (73\%)\end{tabular}} & \multicolumn{3}{c|}{\begin{tabular}[c]{@{}c@{}}2.4\%\\ (22\%)\end{tabular}} & \multicolumn{3}{c|}{\begin{tabular}[c]{@{}c@{}}9\%\\ (81.8\%)\end{tabular}} & \multicolumn{3}{c}{100\%} \\ \hline
\end{tabular}
}\caption{
    Word substitution overlap by word embedding.
    }\label{Exploring_Word_Substitution_Overlap_By_Word_Embedding}
\end{table}

\begin{table}
\centering
\scalebox{0.43}{

\begin{tabular}{lllccccccccc}
\hline
\multicolumn{12}{c}{\textbf{Vocabulary/Dictionary overlap by word embedding types}} \\ \hline
\multicolumn{2}{l}{\multirow{2}{*}{}} & \multicolumn{2}{c}{\multirow{2}{*}{}} & \multicolumn{8}{c}{Word Embedding Overlap} \\ \cline{5-12} 
\multicolumn{2}{l}{} & \multicolumn{2}{c}{} & \multicolumn{2}{c|}{Counter-Fitted} & \multicolumn{2}{c|}{WordNET} & \multicolumn{2}{c|}{Contextualized} & \multicolumn{2}{c}{Glove} \\ \hline
\multicolumn{2}{l}{\begin{tabular}[c]{@{}l@{}}Word \\ Embeddings\end{tabular}} & \multicolumn{2}{l}{\begin{tabular}[c]{@{}l@{}}Number of \\ tokens or \\ words\end{tabular}} & \multicolumn{2}{c|}{65713} & \multicolumn{2}{c|}{147306} & \multicolumn{2}{c|}{30522} & \multicolumn{2}{c}{400000} \\ \hline
\multicolumn{2}{l}{Counter-Fitted} & \multicolumn{2}{l}{65713} & \multicolumn{2}{c|}{65713} & \multicolumn{2}{c|}{\begin{tabular}[c]{@{}c@{}}31233\\ (21.2\% WordNet)\\ (47.5\% Counter)\end{tabular}} & \multicolumn{2}{c|}{\begin{tabular}[c]{@{}c@{}}20994\\ (68.8\% Context)\\ (31.9\% Counter)\end{tabular}} & \multicolumn{2}{c}{\begin{tabular}[c]{@{}c@{}}62951\\ (15.73\% Glove) \\ (95.79\% Counter)\end{tabular}} \\ \hline
\multicolumn{2}{l}{WordNET} & \multicolumn{2}{l}{147306} & \multicolumn{2}{c|}{\begin{tabular}[c]{@{}c@{}}31233\\ (21.2\% WordNet)\\ (47.5\% Counter)\end{tabular}} & \multicolumn{2}{c|}{147306} & \multicolumn{2}{c|}{\begin{tabular}[c]{@{}c@{}}14510\\ (9.85\% WordNet)\\ (47.5\% Context)\end{tabular}} & \multicolumn{2}{c}{\begin{tabular}[c]{@{}c@{}}55666\\ (37.78 WordNet)\\ (13.9 Glove)\end{tabular}} \\ \hline
\multicolumn{2}{l}{Contextualized} & \multicolumn{2}{l}{30522} & \multicolumn{2}{c|}{\begin{tabular}[c]{@{}c@{}}20994\\ (68.8\% Context)\\ (31.9\% Counter)\end{tabular}} & \multicolumn{2}{c|}{\begin{tabular}[c]{@{}c@{}}14510\\ (47.5\% Context)\\ (9.85\% WordNet)\end{tabular}} & \multicolumn{2}{c|}{30522} & \multicolumn{2}{c}{\begin{tabular}[c]{@{}c@{}}22877\\ (75\% Context)\\ (5.7\% Glove)\end{tabular}} \\ \hline
\multicolumn{2}{l}{Glove} & \multicolumn{2}{l}{400000} & \multicolumn{2}{c|}{\begin{tabular}[c]{@{}c@{}}62951\\ (15.73\% Glove)\\ (95.79\% Counter)\end{tabular}} & \multicolumn{2}{c|}{\begin{tabular}[c]{@{}c@{}}55666\\ (13.9 Glove)\\ (37.78 WordNet)\end{tabular}} & \multicolumn{2}{c|}{\begin{tabular}[c]{@{}c@{}}22877\\ (5.7\% Glove)\\ (75\% Context)\end{tabular}} & \multicolumn{2}{c}{400000} \\ \hline
\end{tabular}
}\caption{
    Overlap between vocabulary. Percentage in brackets represents the percentage overlap of the union over the original vocabulary.
    }\label{Vocabulary_Dictionary_Overlap_By_Word_Embedding_Types}
\end{table}

\section{Distance measures}\label{distance_measures}

\subsubsection{MMD}
The general equation for the maximum mean discrepancy (MMD) distance is often described mathematically as the following:

\begin{align}
MMD(P,Q) &= \left\lVert\mathbb{E}_{x\sim P}[\varphi(f(x))]\right. \\
&\quad - \left.\mathbb{E}_{y\sim Q}[\varphi(f(y))]\right\rVert^2
\end{align}

Where $\mathbb{E}$ denotes the expected value, $x$ and $y$ are samples drawn from distributions $P$ and $Q$, respectively, $\varphi$ is a kernel function that maps the data into a high-dimensional feature space, $f_{\theta}$ is an optional feature extractor, and $||.||^2$ is the squared norm of the difference between the two expected values.

In our case, we use $f_{\theta}$ and is set to a sequence classifier. We therefore define $f_{\theta}(x) = h^{A}$ and $f_{\theta}(y) = f_{\theta}(\hat{x}) = h^{B}$. Where in this case both $h^{A}$ and $h^{B}$ are sampled from distributions $H^A$ and $H^B$ respectively. 


\begin{align}
MMD(H^A,H^B) &= \left\lVert\mathbb{E}_{h^{A}\sim H^A}[\varphi(h^{A})]\right. \nonumber \\
&\quad - \left.\mathbb{E}_{h^{B}\sim H^B}[\varphi(h^{B})]\right\rVert^2
\end{align}

We then set $\varphi$ to be a rbf kernel. 


\begin{align}
MMD(H^A,H^B) &= \left\lVert\mathbb{E}_{h^{A}\sim H^A}[h^{A}]\right. \nonumber \\
&\quad - \left.\mathbb{E}_{h^{B}\sim H^B}[h^{B}]\right\rVert^2
\end{align}

\subsubsection{CORAL}
The CORrelation ALignment (CORAL) measures the difference in the second-order statistics of the samples from the two sets. Specifically, it aligns the covariance matrices of the two sets by minimizing the Euclidean norm between them. CORAL requires the covariance matrices to be computable. It can be defined mathematically as follows:

\begin{equation}
\text{CORAL}(H^A,H^B) = \frac{1}{4d^2}\left\lVert C_{H^A} - C_{H^B}\right\rVert_F^2
\end{equation}

 Where $d$ is the dimensionality of the samples, and $C_{H^A}$ and $C_{H^B}$ are the covariance matrices of the samples from ${H^A}$ and ${H^B}$, respectively. The Frobenius/Euclidean norm is denoted by $\left\lVert \cdot \right\rVert_F$. The equation measures the distance between the covariance matrices of ${H^A}$ and ${H^B}$ after aligning them through matrix square root transformation, and scales the distance by a constant factor.

 In the above equation, the covariance matrices are calculated both mathematically and programmatically in the following way:   

 \begin{align}
 C_{H_A} = \frac{1}{n_A-1}(H_A^{T}H_A-\frac{1}{n_A} (1^TH_A)^T(1^TH_A)  )
\end{align}

 \begin{align}
 C_{H_B}  = \frac{1}{n_B-1}(H_B^{T}H_B-\frac{1}{n_B} (1^TH_B)^T(1^TH_B)  )
\end{align}

Where $n_A$/$n_B$ are the number of discrete samples for both the $H_A$/$H_B$ distributions.

 

\subsubsection{Optimal Transport}
There are multiple ways to use optimal transport in the training procedure. One method is first to calculate the p-norm cost matrix $C$ between final representations from the model undergoing base inference $H^B$ and adversarial inference $H^A$ and the transport map $T$. The transport map is the joint probability between $p^{Base}$ and $p^{Adv}$. Where $p^{Base}$ and $p^{Adv}$ are the empirical distributions of the base and adversarial datasets in the feature spaces that satisfy $\sum_{i=1}^{L_B}p^{Base}_{i} = 1$ and $\sum_{j=1}^{L_A}p^{Adv}_{j} = 1$ , often assumed to be uniform and meeting the following conditions.

\begin{equation}
p^{Base}  = \sum_{i=1}^{L_B}p^{Base}_{i}\delta_{h_{i}^{B}} , p^{Adv}  = \sum_{j=1}^{L_A}p^{Adv}_{j}\delta_{h_{j}^{A}}
\end{equation}

Matrix $C$ is later converted to its entropic regularized version with $K = exp(-C/\varepsilon)$. Together with the optimal transport map, the Sinkhorn algorithm \cite{Sinkhorn,Sinkhorn2,Sinkhorn3,Thong} iteratively projects K between $p^{Base}$ and $p^{Adv}$ through multiple Sinkhorn iterations. The distance using Sinkhorn is hence defined as:

\begin{equation}
\begin{split} 
 \mathcal{L}_{Sinkhorn} = D(H^B,H^A) & = OT(p^{Base},p^{Adv}) \\
 & - \frac{1}{2}OT(p^{Base},p^{Base}) \\ 
 & - \frac{1}{2}OT(p^{Adv},p^{Adv})
\end{split}
\end{equation}

Where 

\begin{equation} 
OT(p^{Base},p^{Adv}) = \sum_{i=1}^{M^B}p^{Base}_{i}S^B   +  \sum_{i=1}^{M^A}p^{Adv}_{i}S^A 
\end{equation}

$M^B = M^A$ and $S^A$ and $S^B$ are computed in the aforementioned Sinkhorn iteration. These iterations are further illustrated in Algorithm \ref{algo_sinkhorn}. Sinkhorn iteratively projects the regularised cost matrix K on $p^{Base}$ and $p^{Adv}$.

\begin{algorithm}[]
\caption{Sinkhorn loop}
\label{algo_sinkhorn}
\textbf{Input}:Probabilities $p^{Base}$ and $p^{Adv}$, cost function $C(x_i,y_i)$, Number of iterations N, temperature $\varepsilon$  \\
\textbf{Output}: Sinkhorn tensors $S^A$ and $S^B$ \\
\begin{algorithmic}[1] 
\small
\STATE Initialize $S^A,S^B \leftarrow 0_{R^N}, 0_{R^M}  $
\FOR { Iteration i in N }

    \STATE $S^{A}_{i} \leftarrow -\varepsilon log \sum_{j=1}^{M}p^{Adv}exp\frac{1}{\varepsilon}[S^B_j - C(x_i,y_i)] $
    \STATE  $S^{B}_{j} \leftarrow -\varepsilon log \sum_{i=1}^{N}p^{Base}exp\frac{1}{\varepsilon}[S^A_i - C(x_i,y_i)] $
  
\ENDFOR
\STATE return $S^A,S^B$

\end{algorithmic}
\end{algorithm}

\section{Alternative adversarial training formulations}
There are numerous loss functions that have been demonstrated to enhance the adversarial robustness of various models. However, all of the functions presented in Table \ref{defence_regularizers} have been shown to be ineffective in the language domain according to \cite{FGPM}. In our work, we introduce a new regularizer, which has shown empirical effectiveness on language-related problems.

\begin{table*}[h]
\centering
\scalebox{0.90}
{
\begin{tabular}{|c|c|}
\hline
\textbf{Defense Method} & \textbf{Loss Function} \\
\hline
Standard & $\lambda_{0}\cdot\text{CE}(f_{\text{cls}}(x, \cdot), y) + \lambda_{1}\cdot\text{CE}(f_{\text{cls}}(\hat{x}, \cdot), y)$ \\
\hline
TRADES \cite{TRADES} & $\text{CE}(f_{\text{cls}}(x, \cdot), y) + \lambda \cdot \lVert f_{\text{cls}}(x, \cdot) - f_{\text{cls}}(\hat{x}, \cdot) \rVert_k$ \\
\hline
MMA \cite{MMA} & $\text{CE}(f_{\text{cls}}(x, \cdot), y) \cdot \mathbf{1}(\phi(x) \neq y) + \text{CE}(f_{\text{cls}}(\hat{x}, \cdot), y) \cdot \mathbf{1}(\phi(x) = y)$ \\
\hline
MART \cite{MART} & $\text{BCE}(f_{\text{cls}}(\hat{x}, \cdot), y) + \lambda \cdot \text{KL}(f_{\text{cls}}(x, \cdot) \Vert f_{\text{cls}}(\hat{x}, \cdot)) \cdot (1 - f_{\text{cls}}(x, y))$ \\
\hline
CLP \cite{CLP} & $\text{CE}(f_{\text{cls}}(x, \cdot), y) + \lambda \cdot \lVert f_{\text{cls}}(x, \cdot) - f_{\text{cls}}(x_0, \cdot) \rVert_k$ \\
\hline
SemRoDe (Ours) & $\text{CE}(f_{\text{cls}}(x, \cdot), y) + \lambda \cdot \frac{1}{|\mathcal{B}|} \sum_{\mathcal{B} \subset \mathcal{D}} \text{Dist}(f_{\text{pool}}(t(\mathcal{X}_{B})), f_{\text{pool}}(t(\mathcal{X}_{A})))$ \\
\hline
\end{tabular}
}
\caption{Defense Methods and Loss Functions}\label{defence_regularizers}
\end{table*}

\subsection{Training hyper-parameters}
\label{training_hyper-parameters} Table \ref{parameters_vs_all_plot_over_epochs} outlines the parameters pertaining to Table \ref{vs_all_plot_over_epochs}. The 'Freeze layers' section emphasizes that the initial 11 layers of the models were frozen, serving as a feature extractor. This approach allows us to concentrate on aligning the high-level features during the training process.

\begin{table}[h]
\centering
\scalebox{0.38}{
\begin{tabular}{lllcccccccccccc}
\hline
\multicolumn{1}{c}{\textbf{Model}} & \multicolumn{1}{c}{\textbf{Dataset}} & \multicolumn{1}{c}{\textbf{Method}} & \textbf{\begin{tabular}[c]{@{}c@{}}Base \\ Lambda\end{tabular}} & \textbf{\begin{tabular}[c]{@{}c@{}}Adv \\ Lambda\end{tabular}} & \textbf{\begin{tabular}[c]{@{}c@{}}Dist \\ Lambda\end{tabular}} & \multicolumn{2}{c}{\textbf{\begin{tabular}[c]{@{}c@{}}Data \\ ratio\end{tabular}}} & \multicolumn{2}{c}{\textbf{LR}} & \multicolumn{5}{c}{\textbf{\begin{tabular}[c]{@{}c@{}}Freeze \\ Layers\end{tabular}}} \\ \hline
\multicolumn{1}{c}{\multirow{14}{*}{BERT}} & \multirow{7}{*}{AGNEWS} & Vanilla & 1 & 0 & 0 & \multicolumn{2}{c}{0.1} & \multicolumn{2}{c}{2e-5} & \multicolumn{5}{c}{False} \\
\multicolumn{1}{c}{} &  & TextFooler + Adv Aug & 1 & 1 & 0 & \multicolumn{2}{c}{0.1} & \multicolumn{2}{c}{2e-5} & \multicolumn{5}{c}{False} \\
\multicolumn{1}{c}{} &  & TextFooler + Adv Reg & 1 & 1 & 0 & \multicolumn{2}{c}{0.1} & \multicolumn{2}{c}{2e-5} & \multicolumn{5}{c}{False} \\
\multicolumn{1}{c}{} &  & A2T & 1 & 1 & 0 & \multicolumn{2}{c}{1} & \multicolumn{2}{c}{2e-5} & \multicolumn{5}{c}{False} \\
\multicolumn{1}{c}{} &  & FreeLB++ & 0 & 1 & 0 & \multicolumn{2}{c}{1} & \multicolumn{2}{c}{2e-5} & \multicolumn{5}{c}{False} \\ \cline{3-15} 
\multicolumn{1}{c}{} &  & TextFooler + MMD & 1 & 0 & 1 & \multicolumn{2}{c}{0.1} & \multicolumn{2}{c}{2e-5} & \multicolumn{5}{c}{True}   \\ \cline{2-15} 
\multicolumn{1}{c}{} & \multirow{7}{*}{MR} & Vanilla & 1 & 0 & 0 & \multicolumn{2}{c}{0.1} & \multicolumn{2}{c}{2e-5} & \multicolumn{5}{c}{False} \\
\multicolumn{1}{c}{} &  & TextFooler + Adv Aug & 1 & 1 & 0 & \multicolumn{2}{c}{0.1} & \multicolumn{2}{c}{2e-5} & \multicolumn{5}{c}{False} \\
\multicolumn{1}{c}{} &  & TextFooler + Adv Reg & 1 & 1 & 0 & \multicolumn{2}{c}{0.1} & \multicolumn{2}{c}{2e-5} & \multicolumn{5}{c}{False} \\
\multicolumn{1}{c}{} &  & A2T & 1 & 1 & 0 & \multicolumn{2}{c}{1} & \multicolumn{2}{c}{2e-5} & \multicolumn{5}{c}{False} \\
\multicolumn{1}{c}{} &  & FreeLB++ & 0 & 1 & 0 & \multicolumn{2}{c}{1} & \multicolumn{2}{c}{2e-5} & \multicolumn{5}{c}{False} \\ \cline{3-15} 
\multicolumn{1}{c}{} &  & TextFooler + MMD & 1 & 0 & 1 & \multicolumn{2}{c}{0.1} & \multicolumn{2}{c}{2e-5} & \multicolumn{5}{c}{True}   \\ \hline
\multirow{14}{*}{RoBERTa} & \multirow{7}{*}{AGNEWS} & Vanilla & 1 & 0 & 0 & \multicolumn{2}{c}{0.1} & \multicolumn{2}{c}{2e-5} & \multicolumn{5}{c}{False} \\
 &  & TextFooler + Adv Aug & 1 & 1 & 0 & \multicolumn{2}{c}{0.1} & \multicolumn{2}{c}{2e-5} & \multicolumn{5}{c}{False} \\
 &  & TextFooler + Adv Reg & 1 & 1 & 0 & \multicolumn{2}{c}{0.1} & \multicolumn{2}{c}{2e-5} & \multicolumn{5}{c}{False} \\
 &  & A2T & 1 & 1 & 0 & \multicolumn{2}{c}{1} & \multicolumn{2}{c}{2e-5} & \multicolumn{5}{c}{False} \\
 &  & FreeLB++ & 0 & 1 & 0 & \multicolumn{2}{c}{1} & \multicolumn{2}{c}{2e-5} & \multicolumn{5}{c}{False} \\ \cline{3-15} 
 &  & TextFooler + MMD & 1 & 0 & 1 & \multicolumn{2}{c}{0.1} & \multicolumn{2}{c}{2e-5} & \multicolumn{5}{c}{True}   \\ \cline{2-15} 
 & \multirow{7}{*}{MR} & Vanilla & 1 & 0 & 0 & \multicolumn{2}{c}{0.1} & \multicolumn{2}{c}{2e-5} & \multicolumn{5}{c}{False} \\
 &  & TextFooler + Adv Aug & 1 & 1 & 0 & \multicolumn{2}{c}{0.1} & \multicolumn{2}{c}{2e-5} & \multicolumn{5}{c}{False} \\
 &  & TextFooler + Adv Reg & 1 & 1 & 0 & \multicolumn{2}{c}{0.1} & \multicolumn{2}{c}{2e-5} & \multicolumn{5}{c}{False} \\
 &  & A2T & 1 & 1 & 0 & \multicolumn{2}{c}{1} & \multicolumn{2}{c}{2e-5} & \multicolumn{5}{c}{False} \\
 &  & FreeLB++ & 0 & 1 & 0 & \multicolumn{2}{c}{1} & \multicolumn{2}{c}{2e-5} & \multicolumn{5}{c}{False} \\ \cline{3-15} 
 &  & TextFooler + MMD & 1 & 0 & 1 & \multicolumn{2}{c}{0.1} & \multicolumn{2}{c}{2e-5} & \multicolumn{5}{c}{True} \\  \hline
\end{tabular}
}
\caption{Training parameters for Table \ref{vs_all_plot_over_epochs}}
 \label{parameters_vs_all_plot_over_epochs}
\end{table}

\subsection{Output layer architecture}
For both BERT and RoBERTa, we extract the output features from each token, which are, in our case, the dimension of 128*768, and pool them with $f^{pool}_{\theta}$ along the first dimension. This process results in an output feature vector of shape 768, which encapsulates the information in the whole input. This vector is subsequently used for the $\mathcal{L}_Dist$ regularizer and passed to $f^{cls}_{\theta}$ for classification. The $f^{cls}_{\theta}$ is configured to a single linear layer. We adopted this architecture due to the small dataset sizes used in our experiments.

\subsection{Computational experiments}
The experiments were conducted on a server with 8 Nvidia Tesla v100-sxm2-32gb GPUs. For all our tests, to ensure reproducibility when training or performing adversarial attacks, we set the seed to 765, one of the two seeds used in the TextAttack framework \cite{TextAttack}. For data sampling and shuffling, we set the seed to 42, the same as TextDefender \cite{TextDefender}. We perform each experiment once where we use 500 adversarial samples to evaluate the model's robustness, this is similar to previous work.

\section{Ablation Studies}
The default setting of $N=50$, $\epsilon=0.5$ and $\lambda=1$, to our knowledge, gives the best results.
\subsection{Regularizer $\lambda$ strength}
The $\mathbf{L}_{Dist}$ regularizer's strength has a positive effect on robust accuracy at strengths of 0.5/1, after which it tends to decrease. The robust accuracy for the vanilla model is significantly lower, clocking in at 13.8\% (Table \ref{table:ablation_regularizer}).

\begin{table}[thp]
\centering
\caption{Ablation study for $\lambda$ on $\mathcal{L}_{Dist}$}
\label{table:ablation_regularizer}
\scalebox{0.6}{
\begin{tabular}{lllcccccccccccc}
\hline
 \multicolumn{2}{l}{\multirow{2}{*}{\textbf{Lambda}}} & \multicolumn{13}{c}{\textbf{TextFooler}} \\
\multicolumn{2}{l}{} & \multicolumn{1}{c}{\textbf{CA (↑)}} & \textbf{AUA (↑)} & \multicolumn{11}{c}{\textbf{ASR (↓)}} \\ \hline
\multicolumn{2}{l}{0} & \multicolumn{1}{c}{86.0} & 13.8 & \multicolumn{11}{c}{83.95} \\
\multicolumn{2}{l}{0.1} & \multicolumn{1}{c}{86.0} & 31.8 & \multicolumn{11}{c}{63.02} \\
\multicolumn{2}{l}{0.5} & \multicolumn{1}{c}{85.8} & 39.6 & \multicolumn{11}{c}{53.85} \\
\multicolumn{2}{l}{1} & \multicolumn{1}{c}{85.8} & 39.4 & \multicolumn{11}{c}{54.08} \\
\multicolumn{2}{l}{5} & \multicolumn{1}{c}{85.4} & 34.4 & \multicolumn{11}{c}{59.72} \\
\multicolumn{2}{l}{10} & \multicolumn{1}{c}{85.6} & 31.8 & \multicolumn{11}{c}{62.85} \\ \hline
\end{tabular}
}
\end{table}

\subsection{Attack strength (Attacking)}
In this experiment, the SemRoDe macro model has been trained on TextFooler with $N=50$, but is subsequently attacked with TextFooler values ranging from N=1 to N=150. This is to compare its robust performance with that of the vanilla model.

As the value of N increases, the robustness of the vanilla model decreases. However, upon reaching N=150 - an amount considerably higher than the typically used N=50 - the performance between the two models converges. At this point, the robustness of the vanilla model remains invariant at 12\%, while the SemRoDe macro model stabilizes at 32\%, clearly exhibiting superior robustness with increased N.

\begin{table}[thp]
\centering
\caption{Ablation study on the attacker's number of word substitutions per word $N$}
\label{table:ablation_embeddings}
\scalebox{0.5}{
\begin{tabular}{lllcccccccccccc}
\hline
 \multicolumn{9}{l}{\multirow{2}{*}{\textbf{Attacker N (Epsilon)}}} & \multicolumn{3}{l}{\textbf{Vanilla}} & \multicolumn{3}{l}{\textbf{\begin{tabular}[c]{@{}l@{}}TextFooler + MMD\\ (SemRoDe)\end{tabular}}} \\
\multicolumn{9}{l}{} & \textbf{CA (↑)} & \textbf{AUA (↑)} & \textbf{ASR (↓)} & \textbf{CA (↑)} & \textbf{AUA (↑)} & \textbf{ASR (↓)} \\ \hline
\multicolumn{9}{l}{1} & 86.6 & 75.8 & 12.47 & 85.8 & 75.2 & 12.35 \\
\multicolumn{9}{l}{5} & 86.6 & 44.6 & 48.5 & 85.8 & 56.2 & 34.5 \\
\multicolumn{9}{l}{10} & 86.6 & 33.8 & 60.97 & 85.8 & 51.8 & 39.63 \\
\multicolumn{9}{l}{50} & 86.0 & 13.8 & 83.95 & 85.8 & 39.4 & 54.08 \\
\multicolumn{9}{l}{100} & 86.6 & 12.0 & 86.14 & 85.8 & 32.6 & 62.0 \\
\multicolumn{9}{l}{150} & 86.6 & 12.0 & 86.14 & 85.8 & 32.6 & 62.0 \\ \hline
\end{tabular}
}
\end{table}

\subsection{Ablation $\epsilon$}
The hyperparameter $\epsilon$ determines the maximum angular semantic similarity that a valid adversarial sample must maintain. A small $\epsilon$ results in a high semantic similarity threshold, as defined by the equation $w=1-\frac{\epsilon}{\pi}$. It's important to note that $\epsilon$ is constrained by the bounds (0, $\pi$)

\begin{table}[thp]
\centering
\caption{Ablation study on the attacker's angular similarity $\epsilon$}
\label{table:ablation_epsilon}
\scalebox{0.5}{
\begin{tabular}{lllcccccccccccc}
\hline
\multicolumn{9}{l}{\multirow{2}{*}{\textbf{Attacker $\epsilon$ (Angular Similarity)}}} & \multicolumn{3}{l}{\textbf{Vanilla}} & \multicolumn{3}{l}{\textbf{\begin{tabular}[c]{@{}l@{}}TextFooler + MMD\\ (SemRoDe)\end{tabular}}} \\
\multicolumn{9}{l}{} & \textbf{CA (↑)} & \textbf{AUA (↑)} & \textbf{ASR (↓)} & \textbf{CA (↑)} & \textbf{AUA (↑)} & \textbf{ASR (↓)} \\ \hline
\multicolumn{9}{l}{ $\pi/96$} & 86.6 & 62.2 & 28.18 & 85.8 & 70.4 & 17.95 \\
\multicolumn{9}{l}{$\pi/48$} & 86.6 & 62.2 & 28.18 & 85.8 & 70.4 & 17.95 \\
\multicolumn{9}{l}{$\pi/24$} & 86.6 & 62.2 & 28.18 & 85.8 & 70.4 & 17.95 \\
\multicolumn{9}{l}{$\pi/12$} & 86.6 & 40.4 & 53.35 & 85.8 & 54.2 & 36.83 \\
\multicolumn{9}{l}{$\pi/6$} & 86.6 & 14.0 & 83.83 & 85.8 & 39.0 & 54.55 \\
\multicolumn{9}{l}{$\pi/3$} & 86.6 & 13.6 & 84.3 & 85.8 & 38.4 & 55.24 \\
\multicolumn{9}{l}{$\pi/2$} & 86.6 & 13.6 & 84.3 & 85.8 & 38.4 & 55.24 \\
\multicolumn{9}{l}{$2\cdot\pi/3$} & 86.6 & 13.6 & 84.3 & 85.8 & 38.4 & 55.24 \\
\multicolumn{9}{l}{$5\cdot\pi/6$} & 86.6 & 13.6 & 84.3 & 85.8 & 38.4 & 55.24 \\
\multicolumn{9}{l}{$\pi$} & 86.6 & 13.6 & 84.3 & 85.8 & 38.4 & 55.24 \\ \hline
\end{tabular}
}
\end{table}

\subsection{White-box attacks}

We also investigate the performance of our technique when an attacker uses gradient information to identify word substitutions in a white-box attack strategy, as described in \cite{Hotflip}. Our regularizer, which aligns distributions, enhances robustness, especially when compared to standard adversarial training using the same dataset, as shown in Table \ref{white_box_hotflip}.

\begin{table}[!htp] 
\centering
\caption{SemRoDe's robustness to white-box attacks on BERT-MR}\label{white_box_hotflip}
\scalebox{0.8}{
\begin{tabular}{lccc}
\hline
\textbf{Approach} & \textbf{CA (↑)} & \textbf{AUA (↑)} & \textbf{ASR (↓)} \\
\hline
Baseline          & 86.6            & 55.6             & 35.8            \\
TextFooler + AT   & 85.8            & 58.6             & 31.7            \\
TextFooler + MMD  & 86.0            & 67.6             & 21.4            \\ 
\hline
\end{tabular}
}
\end{table}

\subsection{Adaptive attacks}

This section analyzes the threat model wherein an attacker knows that the model has been trained on SemRoDe, and is therefore robust to word substitutions. In response, the attacker adopts an adaptive attack strategy, such as that described in \cite{AdaptiveAttacks}. We initiate our adaptive attack by posing the question: How might someone compromise the system, given their understanding of our defenses?

To establish our counter-strategy, we commence by: (1) looking for a new objective, through which the optimization process may be effective in generating adversarial samples. Subsequently, (2) we look for an appropriate method to optimize this objective and uncover adversarial examples, lastly (3) we iterate on our new findings to perform better attacks. In the context of NLP classification, our primary goal, which is optimizing for untargeted misclassification, remains the same. Nevertheless, this objective can be changed to that of targeted misclassification if the task has more than two classes. For the purposes of this section's experimentation, we focus on binary classification with the MR dataset.

For the second strategy (2), although our system is not designed for this, a natural choice might be character insertions. To evaluate against these adaptive attacks, we apply the character insertion method described in \cite{TextBugger}, and additionally consider the punctuation insertion techniques put forth in \cite{EmpiricalPunctuationAttacks}. Remarkably, SemRoDe enhances its robustness to character insertion attacks without training on any specific data points (Table \ref{punctuation_insertions}), a finding that potentially reinforces the theory that adversarial attacks originating in the token space reside in a different distribution to that of normal samples.

Proceeding to the next step (3), we repeat the process, now informed by insights from the previous step. Considering that our algorithm remains robust against character insertions, it is possible that the robust adversarial training method we employ can detect the sequence in which words are substituted during an attack. Throughout the training phase, we utilized a greedy search strategy with word replacement to create adversarial examples. This approach enhances the efficiency of greedy search by initially ranking words based on their importance as determined by saliency, attention score, or the impact on classification confidence when removed. The greedy search then prioritizes these words for substitution. To further assess the robustness of our system, we investigate how it performs when the attacker adopts a different optimization strategy, building upon step (2).

For the refined approach (2), we explore a suite of alternative algorithms to traverse the optimization landscape, opting for conventional greedy search, particle swarm optimization \cite{Sememepso}, and beam search instead of greedy search enriched with word replacement (Table \ref{other_optimization_strategies}).

Lastly, we reflect on our findings on how the defence failed, what we learned from it, and potential improvements. We demonstrate that our technique, which aligns distributions, outperforms the use of adversarial training with a regularizer, when an attacker engages in an adaptive attack strategy. However, we do note a significant drop in robust accuracy when changing the optimization strategy. This shows that there may be limitations when generating data through a specific optimization strategy, such as a greedy search paired with word replacement. To enhance performance across various attack optimization strategies, it might be better to identify an optimization method superior to the current greedy search with word replacement. A more comprehensive and diverse range of samples could be achieved by incorporating a wider set of word substitutions, occurring at various positions within the sentence and in a different order. This approach could significantly increase the diversity of the data.

\begin{table}[thp]
\centering
\caption{Initial testing on adaptive adversarial attacks. We explore character and punctuation insertions. The results are on MR and BERT.}
\label{punctuation_insertions}
\scalebox{0.5}{ 
\begin{tabular}{lcccccc}
\hline
\multirow{2}{*}{\textbf{Approach}} & \multicolumn{3}{c}{\textbf{Character Insertions}} & \multicolumn{3}{c}{\textbf{Punctuation Insertions}} \\
& \textbf{CA (↑)} & \textbf{AUA (↑)} & \textbf{ASR (↓)} & \textbf{CA (↑)} & \textbf{AUA (↑)} & \textbf{ASR (↓)} \\ 
\hline
Baseline            & 86.6 & 48.0 & 44.57 & 86.6 & 36.0 & 58.43 \\
TextFooler + AT     & 85.8 & 45.8 & 46.62 & 85.8 & 34.6 & 59.67 \\
TextFooler + MMD    & 86.0 & 58.8 & 31.63 & 86.0 & 51.0 & 40.7  \\
\hline
\end{tabular}
}
\end{table}

\begin{table*}[thp]
\centering
\caption{Second testing on adaptive adversarial attacks. We explore multiple optimization strategies. The results are on MR and BERT.}
\label{other_optimization_strategies}
\scalebox{0.8}{
\begin{tabular}{lccccccccc}
\hline
\textbf{\begin{tabular}[c]{@{}l@{}}Train method \\ (Defense)\end{tabular}} & \multicolumn{3}{c}{\textbf{\begin{tabular}[c]{@{}c@{}}Particle swarm \\ optimization\end{tabular}}} & \multicolumn{3}{c}{\textbf{\begin{tabular}[c]{@{}c@{}}Beam \\ Search\end{tabular}}} & \multicolumn{3}{c}{\textbf{\begin{tabular}[c]{@{}c@{}}Greedy \\ Search\end{tabular}}} \\
 & \textbf{CA (↑)} & \textbf{AUA (↑)} & \textbf{ASR (↓)} & \textbf{CA (↑)} & \textbf{AUA (↑)} & \textbf{ASR (↓)} & \textbf{CA (↑)} & \textbf{AUA (↑)} & \textbf{ASR (↓)} \\ \hline
Baseline & 86.6 & 13.6 & 84.3 & 86.6 & 15.6 & 81.99 & 86.6 & 10.2 & 88.22 \\
TextFooler + AT & 85.8 & 25.6 & 70.16 & 85.8 & 23.6 & 72.49 & 85.8 & 12.6 & 85.31 \\
TextFooler + MMD & 86.0 & 38.6 & 55.12 & 86.0 & 28.2 & 67.21 & 86.0 & 27.4 & 68.14 \\ \hline
\end{tabular}
}
\end{table*}

\section{Extended Computation time}\label{appendix:computation_time}

In this section, we investigate the GPU computation time required for generating the adversarial training set and for training the model with this augmented dataset. Data augmentation (TextFooler + AUG) emerges as the fastest method. Given that the augmented data constitutes only 10\% of the original dataset, the incremental cost during training is minimal. Consequently, data augmentation entails a training duration that is approximately 10\% longer than that of the baseline model.

The Adversarial Training techniques TextFooler + AT, A2T, and TextFooler + MMD (SemRoDe), work by first creating an adversarial training set. The lower computation times between TextFooler and A2T show that data generation with TextFooler is more computationally efficient than with A2T. Meanwhile, with respect to the training process, these methods exhibit similar timeframes, approximately doubling the duration compared to baseline training. This increase is attributable to the necessity of sampling and conducting inference using the adversarial dataset at each base training iteration.

We also considered embedding perturbation techniques such as FreeLB++, InfoBERT and DSRM. These approaches involve modifying the input through multiple PGD steps at every training iteration. To match the performance of the baseline, as detailed in the implementation from TextDefender \cite{TextDefender}. Specifically, we applied the gradient perturbation 15 times for InfoBERT, 30 times for FreeLB++, and once for DSRM per iteration. The accumulation of these multiple PGD steps across a comprehensive dataset significantly extends the training duration. Table \ref{table:computation_cost} provides a comparative analysis of the computational times of the different methods.




\section{Distinction between GloVe and Counter-Fitted GloVe embeddings}

The word embeddings used in this study are GloVe embeddings. GloVe \cite{GloVe} embeddings are generated using a self-supervised algorithm that extracts global word-word co-occurrence statistics from a given corpus. However, to address certain limitations of plain GloVe embeddings, new authors introduced Counter-Fitted GloVe embeddings \cite{Counter_Fitted_Embeddings}. These counter-fitted embeddings aim to mitigate issues arising from the expectation of semantic similarity, which sometimes results in conceptual associations and anomalies within the generated embedding clusters. These anomalies may include clusters containing antonyms instead of synonyms, or dissimilar words that typically appear in similar contexts, such as east/west/north/south.

Counter-fitted embeddings inject both synonym and anonym information into the original GloVe embeddings, making them more suitable for word substitution attacks. In the context of this research, the original TextBugger paper employed GloVe embeddings, but within the TextAttack framework, the implementation of TextBugger utilizes Counter-Fitted GloVe embeddings. The experiments conducted for this study were performed using the original TextBugger algorithm with GloVe embeddings as the baseline for comparison, while TextFooler utilizes the Counter-Fitted GloVe embeddings.

\section{Extended Feature space plots}\label{appendix:extended_plots}
\begin{figure}[h!]
    \centering
    \includegraphics[width=.45\textwidth]{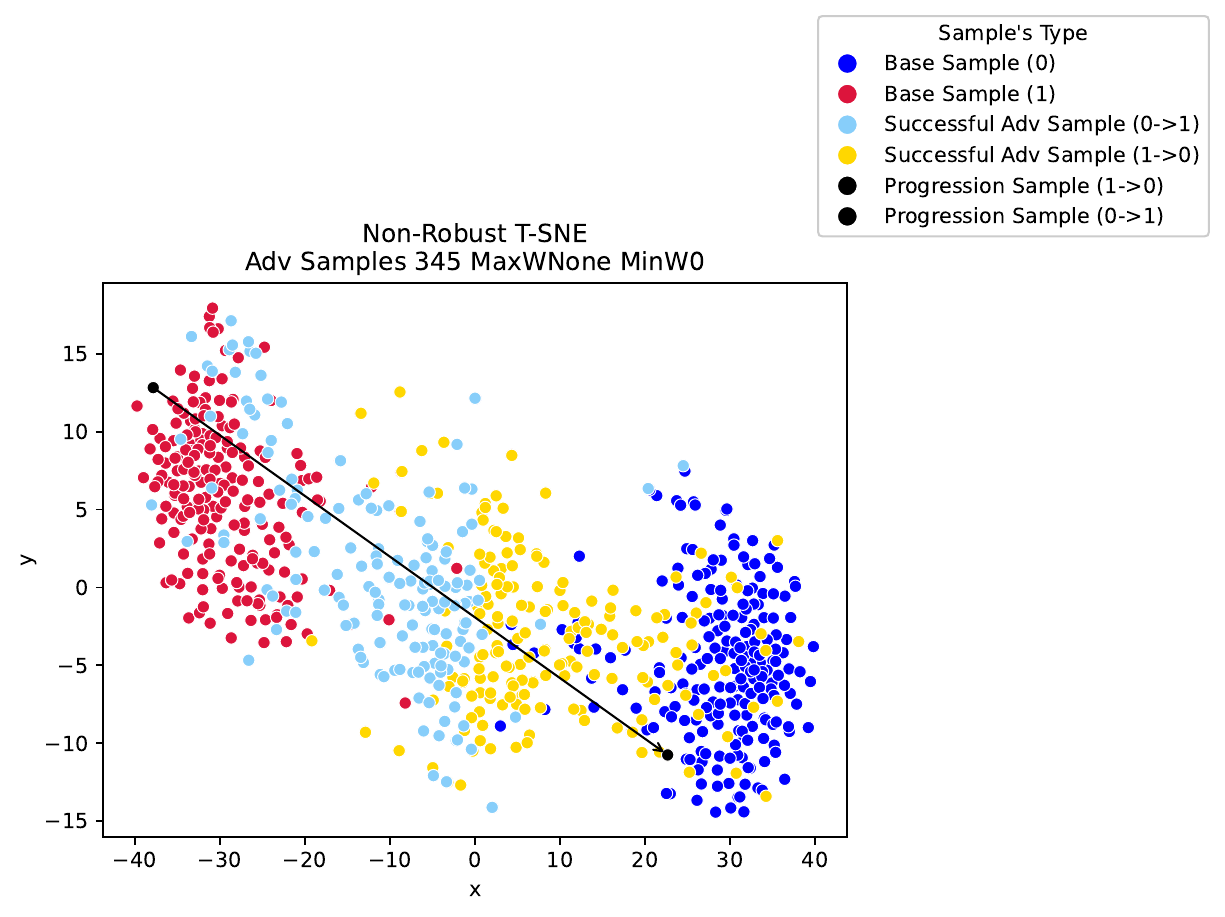}  
 \caption{
     Non-Robust model TSNE.  
    }\label{img:TSNE_OT_Non_Robust74}
\end{figure}

\begin{figure}[h!]
    \centering
    \includegraphics[width=.45\textwidth]{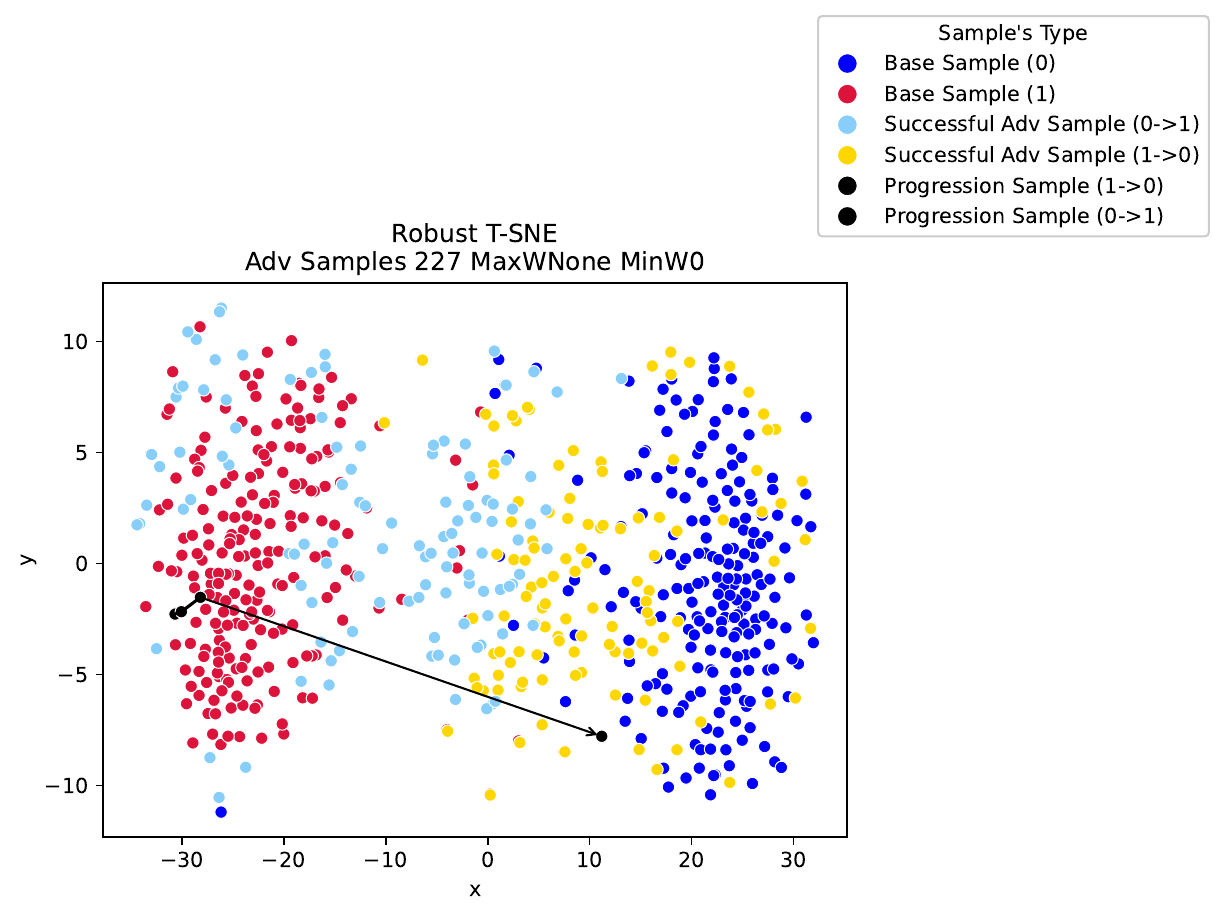}  
 \caption{
     Robust model TSNE.  
    }\label{img:TSNE_OT_Robust74}
\end{figure}

\begin{figure}[h!]
    \centering
    \includegraphics[width=.45\textwidth]{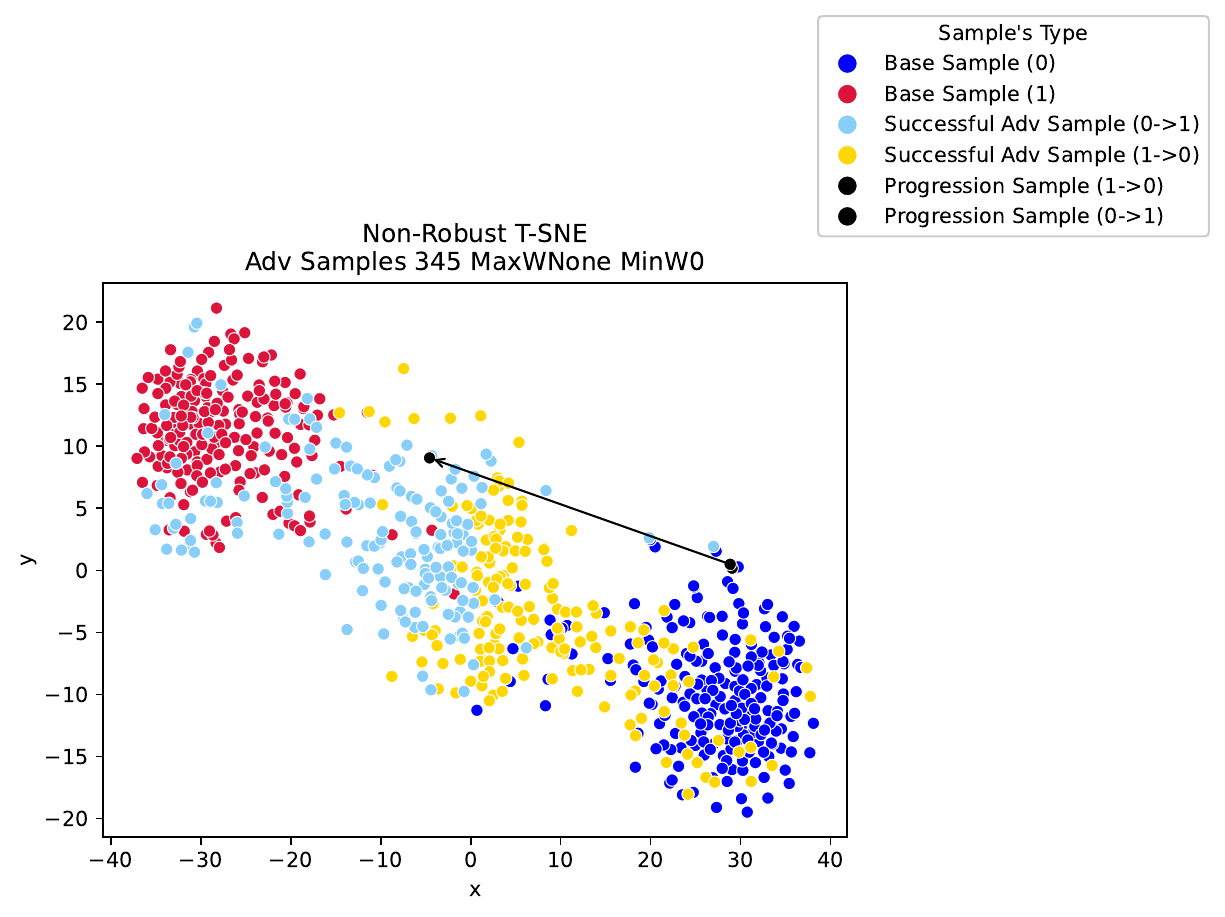}  
 \caption{
     Non-Robust model TSNE.  
    }\label{img:TSNE_OT_Non_Robust_401}
\end{figure}

\begin{figure}[h!]
    \centering
    \includegraphics[width=.45\textwidth]{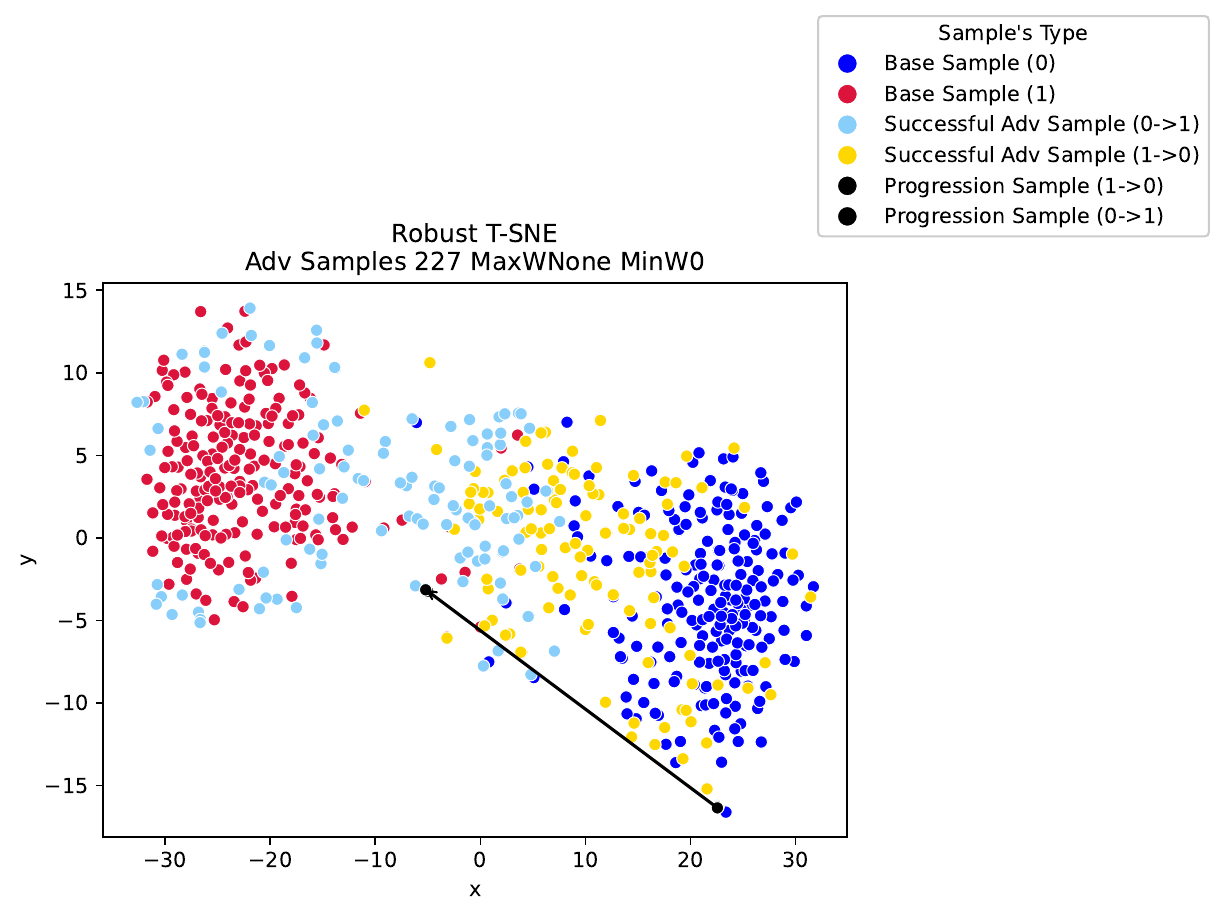}  
 \caption{
     Robust model TSNE.  
    }\label{img:TSNE_OT_Robust_401}
\end{figure}

\begin{figure}[h!]
    \centering
     
    \includegraphics[width=.45\textwidth]{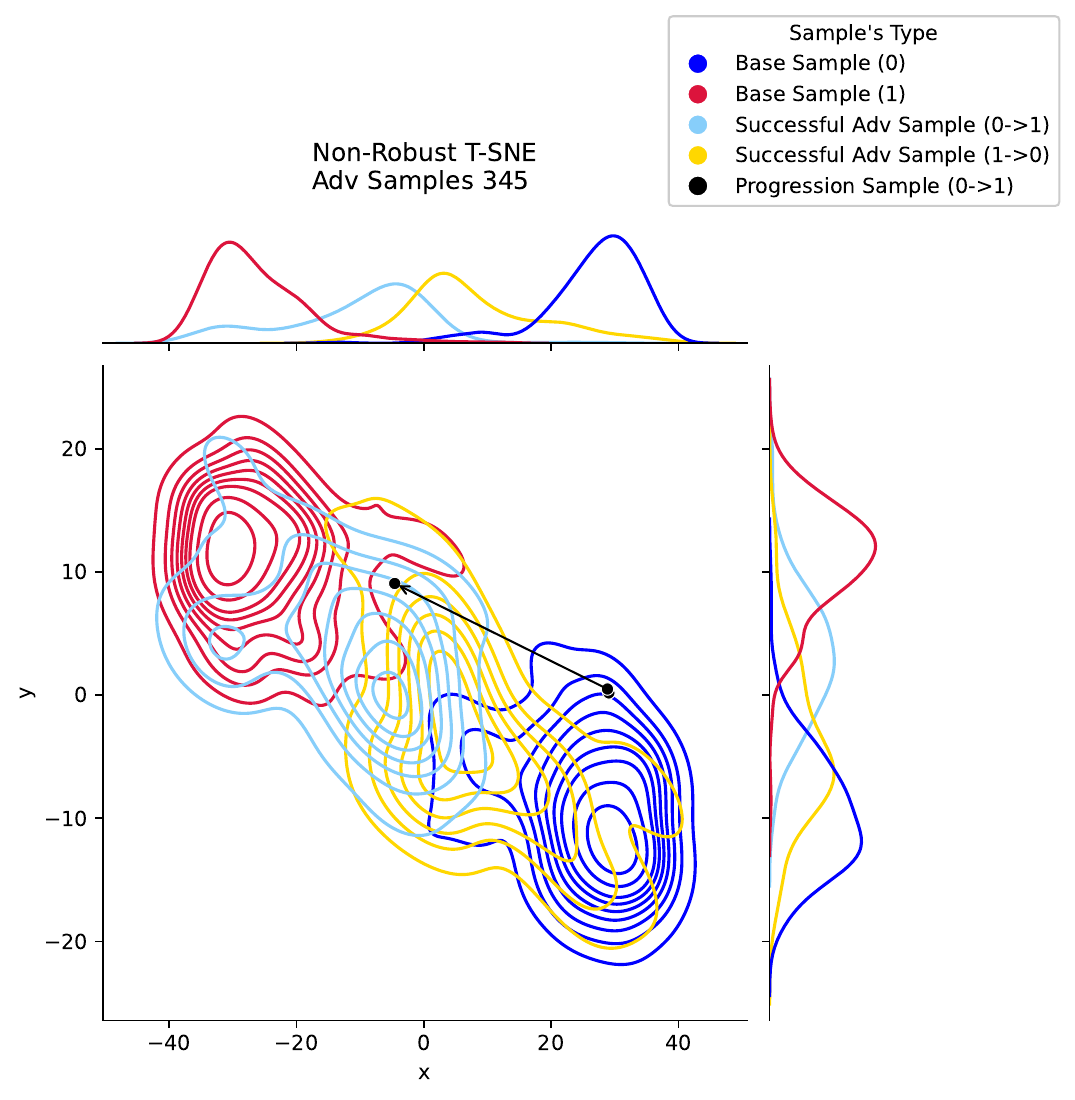} 
 \caption{
     Non Robust model TSNE with marginal distributions.
    }\label{img:TSNE_OT_Non_Robust_MD_401}
\end{figure}

\begin{figure}[h!]
    \centering
     
    \includegraphics[width=.45\textwidth]{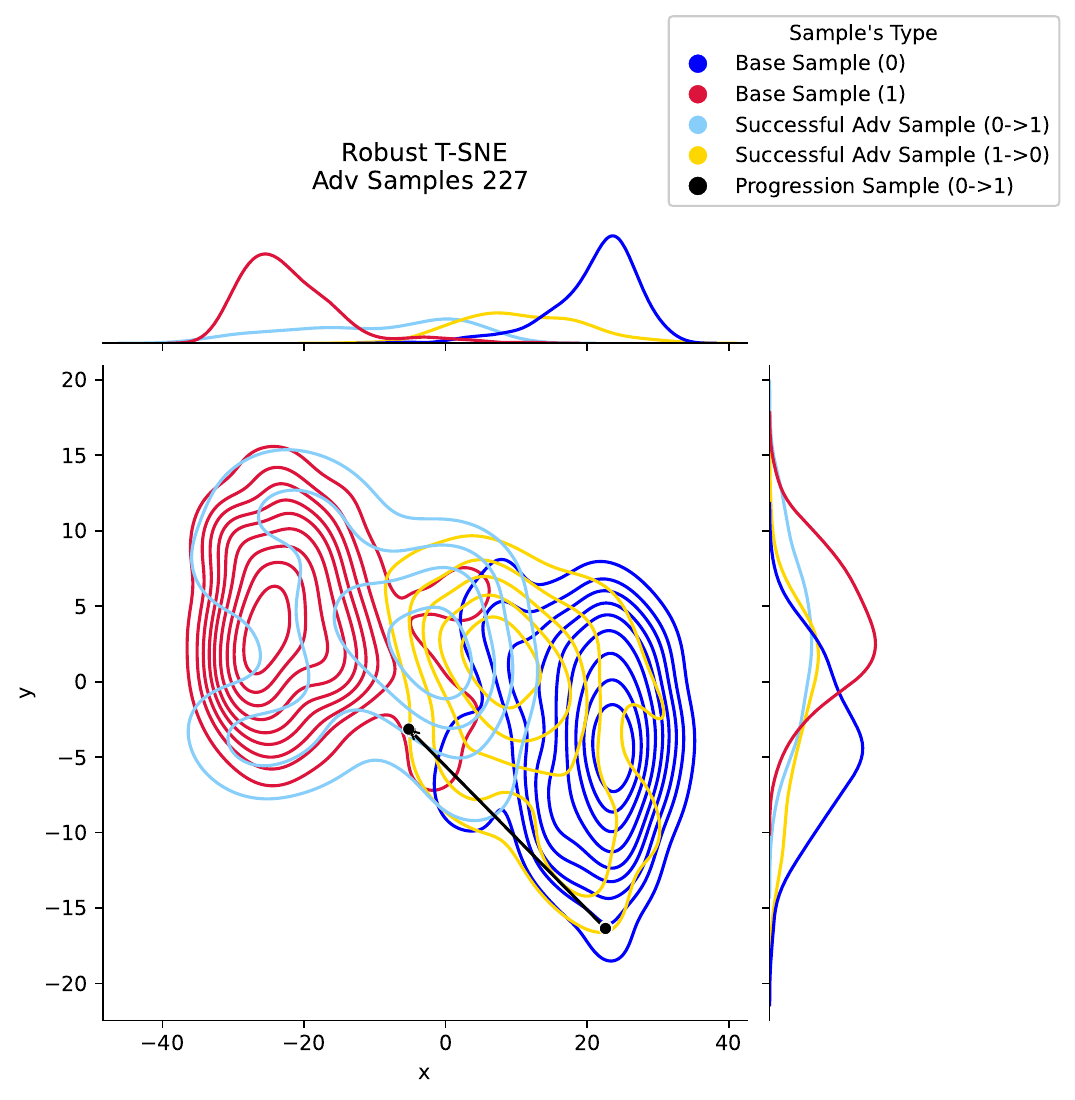} 
 \caption{
     Robust model TSNE with marginal distributions.
    }\label{img:TSNE_OT_Robust_MD_401}
\end{figure}

\begin{figure}[thp]
    \centering
    \includegraphics[width=.85\linewidth]{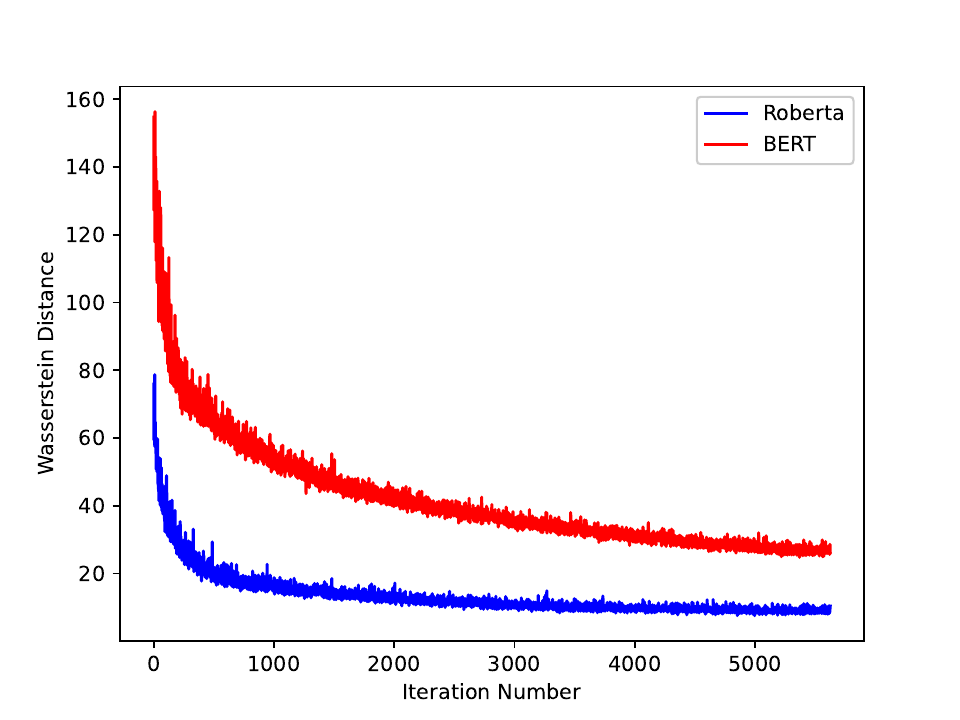} 
    \includegraphics[width=.85\linewidth]{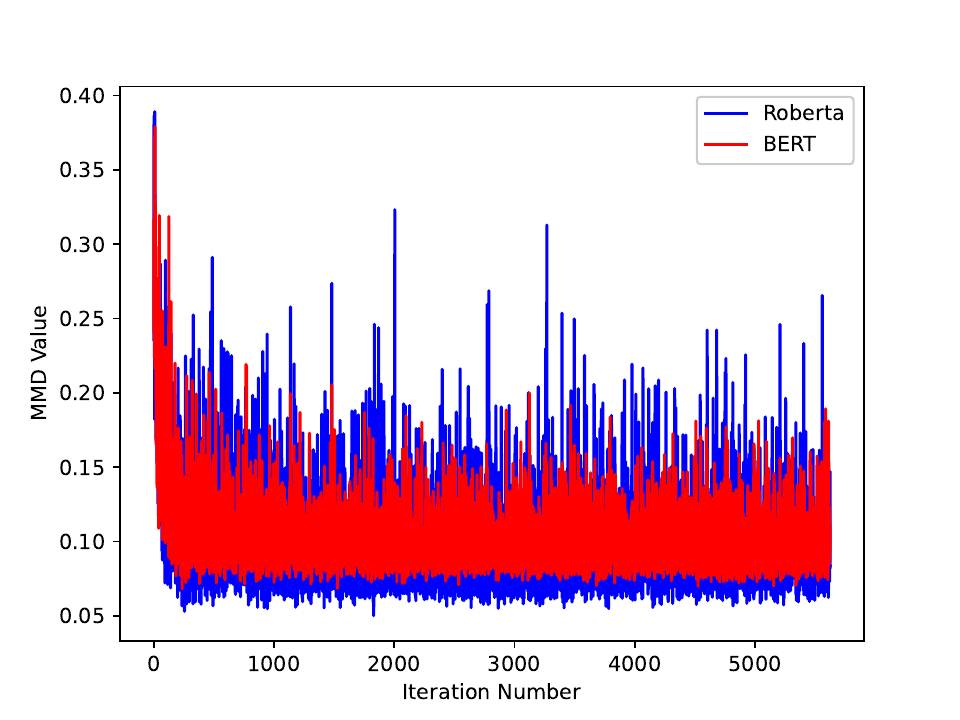} 
    \caption{
         OT (Top) MMD (Bottom) response over iterations for AGNEWS
    }
    \label{img:distance_comparison_plot_AGNEWS}
\end{figure}

\begin{figure}[thp]
    \centering
    \includegraphics[width=.85\linewidth]{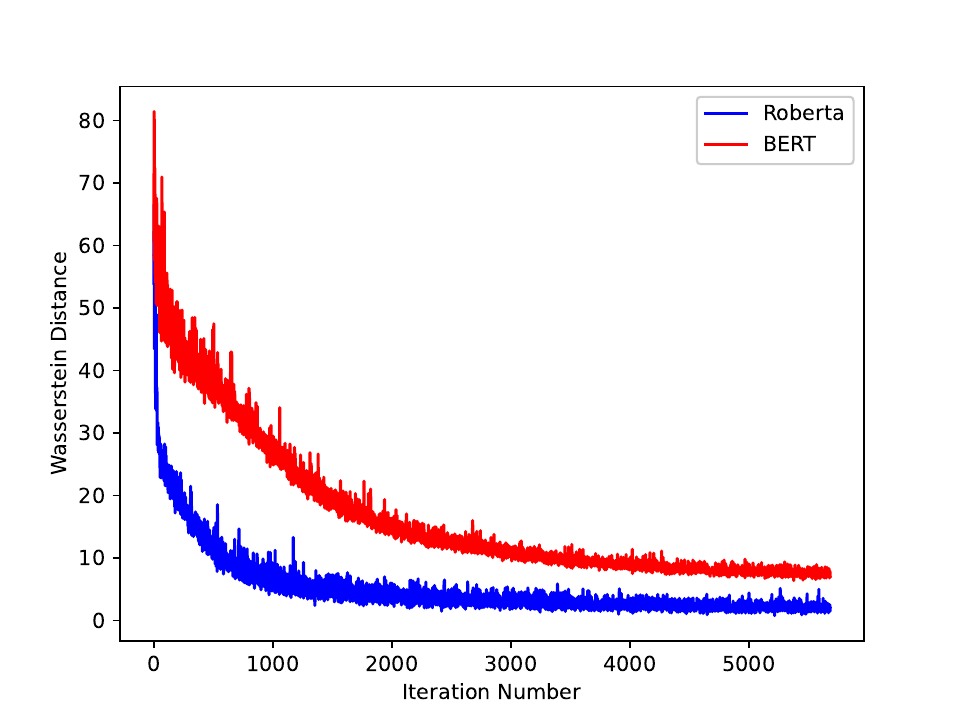} 
    \includegraphics[width=.85\linewidth]{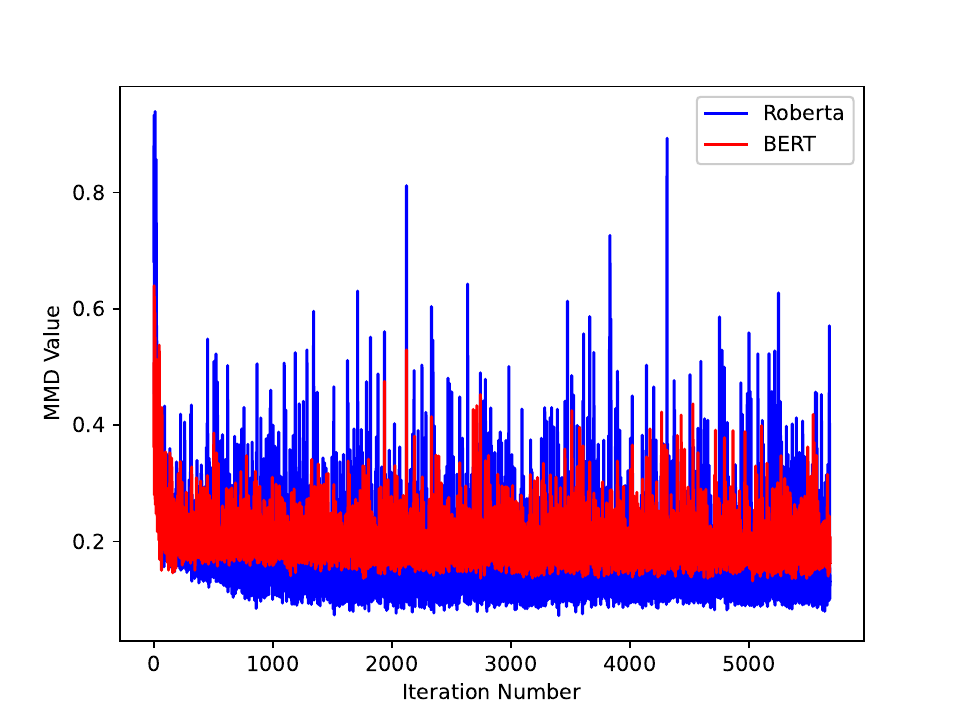} 
    \caption{
         OT (Top) MMD (Bottom) response over iterations for SST2
    }
    \label{img:distance_comparison_plot_SST2}
\end{figure}

Figures \ref{img:TSNE_OT_Non_Robust74}, \ref{img:TSNE_OT_Robust74} present the t-SNE visualizations of MR without marginal distributions, with the corresponding plots showcased in Figure \ref{img:aug_data_distribution}. Meanwhile, Figures \ref{img:TSNE_OT_Non_Robust_401}, \ref{img:TSNE_OT_Robust_401}, \ref{img:TSNE_OT_Non_Robust_MD_401}, \ref{img:TSNE_OT_Robust_MD_401} display the t-SNE and marginal distributions for AG-News. Figure \ref{img:distance_comparison_plot_AGNEWS} illustrates the reduction in distances for AG-News, and Figure \ref{img:distance_comparison_plot_SST2} depicts the decreasing distances for SST2.

\section{Qualitative Samples }
Table \ref{Appendix:qualitative_examples} showcases two examples that are progressively perturbed through word substitutions performed by TextFooler.
\begin{table*}[]
\centering
\scalebox{0.50}{
\begin{tabular}{lllcccccccccccc}
\hline
\multicolumn{1}{c}{\textbf{\textbf{Sample Type}}} & \multicolumn{1}{c}{\textbf{Model}} & \multicolumn{1}{c}{\textbf{\begin{tabular}[c]{@{}c@{}}Number of Word\\ Substitutions\end{tabular}}} & \multicolumn{12}{c}{\textbf{Sample (MR Sentiment classification)}} \\ \hline
\multicolumn{15}{c}{\textbf{Positive Sentiment to Negative Sentiment}} \\ \hline
\multirow{2}{*}{BERT Base} & Base Sample (1) & \multicolumn{1}{c}{0} & \multicolumn{12}{l}{buy is an accomplished actress, and this is a big, juicy role .} \\
 & Successful Adv Sample (1 to 0) & \multicolumn{1}{c}{1} & \multicolumn{12}{l}{buy is an accomplished actress, and this is a big, \textbf{fecund} role .} \\ \cline{2-15} 
\multirow{4}{*}{BERT Robust} & Base Sample (1) & \multicolumn{1}{c}{0} & \multicolumn{12}{l}{buy is an accomplished actress, and this is a big, juicy role .} \\
 & Base Sample (1) & \multicolumn{1}{c}{1} & \multicolumn{12}{l}{\textbf{absorbing} is an accomplished actress, and this is a big, juicy role .} \\
 & Base Sample (1) & \multicolumn{1}{c}{2} & \multicolumn{12}{l}{\textbf{absorbing} is an accomplished actress, and this is a \textbf{momentous}, juicy role .} \\
 & Successful Adv Sample (1 to 0) & \multicolumn{1}{c}{3} & \multicolumn{12}{l}{\textbf{absorbing} is an accomplished actress, and this is a \textbf{momentous}, juicy \textbf{liability} .} \\ \hline
\multicolumn{15}{c}{\textbf{Negative Sentiment to Positive Sentiment}} \\ \hline
\multirow{3}{*}{BERT Base} & Base Sample (0) & \multicolumn{1}{c}{0} & \multicolumn{12}{l}{\begin{tabular}[c]{@{}l@{}}you'll just have your head in your hands wondering why lee's character didn't just go to a \\ bank manager and save everyone the misery .\end{tabular}} \\
 & Base Sample (0) & \multicolumn{1}{c}{1} & \multicolumn{12}{l}{\begin{tabular}[c]{@{}l@{}}you'll just have your head in your \textbf{veins} wondering why lee's character didn't just go to a \\ bank manager and save everyone the misery .\end{tabular}} \\
 & Successful Adv Sample (0 to 1) & \multicolumn{1}{c}{2} & \multicolumn{12}{l}{\begin{tabular}[c]{@{}l@{}}you'll just \textbf{obtain} your head in your \textbf{veins} wondering why lee's character didn't just go to a \\ bank manager and save everyone the misery .\end{tabular}} \\ \cline{2-15} 
\multirow{6}{*}{BERT Robust} & Base Sample (0) & \multicolumn{1}{c}{0} & \multicolumn{12}{l}{\begin{tabular}[c]{@{}l@{}}you'll just have your head in your hands wondering why lee's character didn't just go to a \\ bank manager and save everyone the misery .\end{tabular}} \\
 & Base Sample (0) & \multicolumn{1}{c}{1} & \multicolumn{12}{l}{\begin{tabular}[c]{@{}l@{}}you'll just have your head in your hands wondering why lee's character didn't just go to a \\ bank \textbf{warden} and save everyone the misery .\end{tabular}} \\
 & Base Sample (0) & \multicolumn{1}{c}{2} & \multicolumn{12}{l}{\begin{tabular}[c]{@{}l@{}}you'll just have your head in your hands wondering why lee's character didn't just go to a \\ \textbf{southwest warden} and save everyone the misery .\end{tabular}} \\
 & Base Sample (0) & \multicolumn{1}{c}{3} & \multicolumn{12}{l}{\begin{tabular}[c]{@{}l@{}}you'll just have your head in your hands wondering why lee's \textbf{idiosyncrasies} didn't just go to a \\ \textbf{southwest warden} and save everyone the misery .\end{tabular}} \\
 & Base Sample (0) & \multicolumn{1}{c}{4} & \multicolumn{12}{l}{\begin{tabular}[c]{@{}l@{}}you'll just have your head in your \textbf{veins} wondering why lee's \textbf{idiosyncrasies} didn't just go to a \\ \textbf{southwest warden} and save everyone the misery .\end{tabular}} \\
 & Successful Adv Sample (0 to 1) & \multicolumn{1}{c}{5} & \multicolumn{12}{l}{\begin{tabular}[c]{@{}l@{}}you'll just have your head in your \textbf{veins} wondering why lee's \textbf{idiosyncrasies} didn't just go to a \\ \textbf{southwest warden} and save everyone the \textbf{miseries} .\end{tabular}} \\ \hline
\end{tabular}
}\caption{
    Qualitative examples showing a sample being perturbed in both the base model and our robust model. Positive to Negative Sentiment (Figures \ref{img:aug_data_distribution} (Top and Bottom), \ref{img:TSNE_OT_Non_Robust74} and \ref{img:TSNE_OT_Robust74}). Negative to Positive Sentiment (Figures \ref{img:TSNE_OT_Non_Robust_401},\ref{img:TSNE_OT_Non_Robust_MD_401}, \ref{img:TSNE_OT_Robust_401}, \ref{img:TSNE_OT_Robust_MD_401}).
    }\label{Appendix:qualitative_examples}
\end{table*}

\section{Extended SemRoDe results}\label{Extended_MAT_results}
We also test on RoBERTAa and AGNEWS in Table \ref{Appendix:distribution_alignment_mmd}.

\begin{table*}[]
\centering
\scalebox{0.50}{

\begin{tabular}{lllcccccccccccc}
\hline
\multirow{3}{*}{\textbf{Model}} & \multirow{3}{*}{\textbf{Dataset}} & \multirow{3}{*}{\textbf{\begin{tabular}[c]{@{}l@{}}Train method \\ (Defense)\end{tabular}}} & \multicolumn{12}{c}{\textbf{Test Method}} \\
 &  &  & \multicolumn{3}{c}{\textbf{\begin{tabular}[c]{@{}c@{}}PWWS  \\ (WordNet)\end{tabular}}} & \multicolumn{3}{c}{\textbf{\begin{tabular}[c]{@{}c@{}}BERTAttack  \\ (Contextual)\end{tabular}}} & \multicolumn{3}{c}{\textbf{\begin{tabular}[c]{@{}c@{}}TextFooler \\ (Counter Fitted)\end{tabular}}} & \multicolumn{3}{c}{\textbf{\begin{tabular}[c]{@{}c@{}}TextBugger\\ (Sub-W GloVe)\end{tabular}}} \\
 &  &  & \textbf{CA (↑)} & \textbf{AUA (↑)} & \textbf{ASR (↓)} & \textbf{CA (↑)} & \textbf{AUA (↑)} & \textbf{ASR (↓)} & \textbf{CA (↑)} & \textbf{AUA (↑)} & \textbf{ASR (↓)} & \textbf{CA (↑)} & \textbf{AUA (↑)} & \textbf{ASR (↓)} \\ \hline
\multirow{33}{*}{\textbf{BERT}} & \multirow{11}{*}{\textbf{AGNews}} & Vanilla & 96.2 & 56.2 & 41.58 & 96.2 & 45.8 & 52.39 & 96.2 & 30.4 & 68.4 & 96.0 & 41.4 & 56.88 \\
 &  & TextFooler + Adv Aug & 95.4 & 66.4 & 30.4 & 95.4 & 49.6 & 48.01 & 95.4 & 51.6 & 45.91 & 95.4 & 50.6 & 46.96 \\
 &  & TextFooler + Adv Reg & 96.4 & 71.2 & 26.14 & 96.4 & {[}55.4{]} & 42.53 & 96.4 & 58.4 & 39.42 & 96.4 & 56.4 & 41.49 \\
 &  & A2T & 96.4 & 58.0 & 39.83 & 96.4 & 45.2 & 53.11 & 96.4 & 41.6 & 56.85 & 96.4 & 43.6 & 54.77 \\
 &  & InfoBERT & 96.0 & 69.8 & 27.29 & 96.0 & 59.6 & 37.92 & 96.0 & 49.2 & 48.75 & 96.0 & 55.8 & 41.88 \\
 &  & FreeLB++ & 95.4 & 73.8 & 22.64 & 95.4 & 54.8 & 42.56 & 95.4 & 52.8 & 44.65 & 95.4 & 51.6 & 45.91 \\
 &  & DSRM & 94.6 & 55.4 & 41.44 & 94.6 & 48.8 & 48.41 & 94.6 & 37.4 & 60.47 & 94.6 & 47.6 & 49.68 \\ \cline{3-15} 
 &  & PWWS + MMD (SemRoDe) & 96.2 & 45.0 & 53.22 & 96.2 & 44.4 & 53.85 & 96.2 & 35.0 & 63.62 & 95.6 & 43.2 & 54.81 \\
 &  & BERTAttack + MMD (SemRoDe) & 95.8 & {[}75.4{]} & 21.29 & 95.8 & \textbf{73.0} & 23.8 & 95.6 & (68.0) & 28.87 & 95.6 & (73.2) & 23.43 \\
 &  & TextBugger + MMD (SemRoDe) & 95.6 & (75.6) & 20.92 & 95.6 & \textbf{73.0} & 23.64 & 95.6 & {[}67.8{]} & 29.08 & 95.6 & {[}73.0{]} & 23.64 \\
 &  & TextFooler + MMD (SemRoDe) & 95.6 & \textbf{77.0} & 19.46 & 95.6 & (72.6) & 24.06 & 95.6 & \textbf{69.4} & 27.41 & 95.6 & \textbf{74.2} & 22.38 \\ \cline{2-15} 
 & \multirow{11}{*}{\textbf{SST-2}} & Vanilla & 94.0 & 20.2 & 78.51 & 94.0 & 31.6 & 66.38 & 94.0 & 11.4 & 87.87 & 94.0 & 2.8 & 97.02 \\
 &  & TextFooler + Adv Aug & 94.0 & 27.4 & 70.85 & 94.0 & 33.0 & 64.89 & 94.0 & 18.4 & 80.43 & 94.0 & 4.2 & 95.53 \\
 &  & TextFooler + Adv Reg & 94.6 & 23.4 & 75.26 & 94.6 & 33.0 & 65.12 & 94.6 & 18.4 & 80.55 & 94.6 & 4.6 & 95.14 \\
 &  & A2T & 94.0 & 22.8 & 75.74 & 94.0 & 38.8 & 58.72 & 94.0 & 16.0 & 82.98 & 94.0 & 5.0 & 94.68 \\
 &  & InfoBERT & 92.6 & 27.4 & 70.41 & 92.6 & 32.2 & 65.23 & 92.6 & 15.8 & 82.94 & 92.6 & 4.0 & 95.68 \\
 &  & FreeLB++ & 90.8 & 28.8 & 68.28 & 90.8 & 35.0 & 61.45 & 90.8 & 16.8 & 81.5 & 90.8 & 4.4 & 95.15 \\
 &  & DSRM & 93.0 & 23.8 & 74.41 & 93.0 & 41.6 & 55.27 & 93.0 & 27.4 & 70.54 & 93.0 & 11.0 & 88.17 \\ \cline{3-15} 
 &  & PWWS + MMD (SemRoDe) & 93.8 & (41.6) & 55.65 & 93.8 & \textbf{59.2} & 36.89 & 93.8 & (43.8) & 53.3 & 93.8 & (30.2) & 67.8 \\
 &  & BERTAttack + MMD (SemRoDe) & 94.2 & \textbf{43.6} & 53.72 & 94.2 & (58.2) & 38.22 & 94.2 & \textbf{49.0} & 47.98 & 94.2 & \textbf{33.2} & 64.76 \\
 &  & TextBugger + MMD (SemRoDe) & 94.2 & 31.4 & 66.67 & 94.2 & 48.8 & 48.2 & 94.2 & 32.6 & 65.39 & 93.8 & 18.6 & 80.17 \\
 &  & TextFooler + MMD (SemRoDe) & 94.2 & {[}40.4{]} & 57.11 & 94.2 & {[}55.4{]} & 41.19 & 94.2 & {[}40.2{]} & 57.32 & 94.2 & {[}25.0{]} & 73.46 \\ \cline{2-15} 
 & \multirow{11}{*}{\textbf{MR}} & Vanilla & 86.0 & 21.2 & 75.35 & 86.0 & 32.6 & 62.09 & 86.0 & 13.8 & 83.95 & 86.0 & 5.8 & 93.26 \\
 &  & TextFooler + Adv Aug & 85.6 & 21.0 & 75.47 & 85.6 & 31.6 & 63.08 & 85.6 & 11.8 & 86.21 & 85.6 & 4.0 & 95.33 \\
 &  & TextFooler + Adv Reg & 86.8 & 25.8 & 70.28 & 86.8 & 34.2 & 60.6 & 86.8 & 18.4 & 78.8 & 86.4 & 8.0 & 90.74 \\
 &  & A2T & 86.8 & 18.0 & 79.26 & 86.8 & 32.6 & 62.44 & 86.8 & 13.8 & 84.1 & 86.8 & 3.8 & 95.62 \\
 &  & InfoBERT & 82.6 & 31.4 & 61.99 & 82.6 & 34.6 & 58.11 & 82.6 & 19.6 & 76.27 & 82.6 & 6.8 & 91.77 \\
 &  & FreeLB++ & 84.2 & 26.6 & 68.41 & 84.2 & 32.4 & 61.52 & 84.2 & 16.4 & 80.52 & 84.2 & 3.8 & 95.49 \\
 &  & DSRM & 87.6 & 25.6 & 70.78 & 87.6 & 32.6 & 62.79 & 87.6 & 20.0 & 77.17 & 87.6 & 11.0 & 87.44 \\ \cline{3-15} 
 &  & PWWS + MMD (SemRoDe) & 86.0 & 37.0 & 56.98 & 86.0 & {[}45.8{]} & 46.74 & 86.0 & 38.2 & 55.58 & 86.0 & 28.2 & 67.21 \\
 &  & BERTAttack + MMD (SemRoDe) & 85.8 & \textbf{38.2} & 55.48 & 85.8 & (46.6) & 45.69 & 85.8 & \textbf{40.0} & 53.38 & 85.8 & \textbf{29.4} & 65.73 \\
 &  & TextBugger + MMD (SemRoDe) & 85.8 & {[}37.4{]} & 56.41 & 85.8 & \textbf{47.6} & 44.52 & 85.8 & (39.4) & 54.08 & 85.8 & (28.8) & 66.43 \\
 &  & TextFooler + MMD (SemRoDe) & 85.8 & (37.8) & 55.94 & 85.8 & 44.6 & 48.02 & 85.8 & {[}39.0{]} & 54.55 & 85.8 & {[}28.6{]} & 66.67 \\ \hline
\multirow{33}{*}{\textbf{RoBERTa}} & \multirow{11}{*}{\textbf{AGNews}} & Vanilla & 95.8 & 49.8 & 48.02 & 95.8 & 45.6 & 52.4 & 95.8 & 32.2 & 66.39 & 95.8 & 44.8 & 53.24 \\
 &  & TextFooler + Adv Aug & 95.4 & 63.0 & 33.96 & 95.4 & 51.4 & 46.12 & 95.4 & 50.6 & 46.96 & 95.4 & 51.8 & 45.7 \\
 &  & TextFooler + Adv Reg & 95.4 & 67.2 & 29.56 & 95.4 & 57.6 & 39.62 & 95.4 & 60.0 & 37.11 & 95.4 & 60.0 & 37.11 \\
 &  & A2T & 94.8 & 61.2 & 35.44 & 94.8 & 49.8 & 47.47 & 94.8 & 46.8 & 50.63 & 94.8 & 50.8 & 46.41 \\
 &  & InfoBERT & 95.6 & 63.4 & 33.68 & 95.6 & 62.8 & 34.31 & 95.6 & 41.0 & 57.11 & 95.6 & 55.6 & 41.84 \\
 &  & FreeLB++ & 96.0 & 70.0 & 27.08 & 96.0 & 64.8 & 32.5 & 96.0 & 46.4 & 51.67 & 96.0 & 59.6 & 37.92 \\
 &  & DSRM & 95.8 & 64.6 & 32.71 & 96.0 & 57.2 & 40.42 & 96.0 & 40.2 & 58.12 & 96.0 & 53.4 & 44.38 \\ \cline{3-15} 
 &  & PWWS + MMD (SemRoDe) & 95.4 & (72.4) & 24.11 & 95.4 & 70.6 & 26.0 & 95.4 & (63.0) & 33.96 & 95.4 & (70.0) & 26.62 \\
 &  & BERTAttack + MMD (SemRoDe) & 95.4 & \textbf{74.8} & 21.59 & 95.4 & \textbf{72.8} & 23.69 & 95.4 & \textbf{70.8} & 25.79 & 95.4 & 61.6 & 35.43 \\
 &  & TextBugger + MMD (SemRoDe) & 95.4 & 71.2 & 25.37 & 95.4 & (71.6) & 24.95 & 95.4 & {[}61.4{]} & 35.64 & 95.4 & {[}69.4{]} & 27.25 \\
 &  & TextFooler + MMD (SemRoDe) & 95.6 & {[}71.4{]} & 25.31 & 95.6 & {[}70.8{]} & 25.94 & 95.6 & 61.2 & 35.98 & 95.6 & \textbf{70.0} & 26.78 \\ \cline{2-15} 
 & \multirow{11}{*}{\textbf{SST2}} & Vanilla & 94.2 & 25.0 & 73.46 & 94.2 & 38.4 & 59.24 & 94.2 & 14.2 & 84.93 & 94.2 & 6.6 & 92.99 \\
 &  & TextFooler + Aug & 93.0 & 35.8 & 61.51 & 93.0 & 38.2 & 58.92 & 93.0 & 29.2 & 68.6 & 93.0 & 12.4 & 86.67 \\
 &  & TextFooler + AT & 93.8 & 30.6 & 67.38 & 93.8 & 34.8 & 62.9 & 93.8 & 21.6 & 76.97 & 93.8 & 4.8 & 94.88 \\
 &  & A2T & 93.6 & 20.6 & 77.99 & 93.6 & 32.8 & 64.96 & 93.6 & 14.0 & 85.04 & 93.6 & 3.6 & 96.15 \\
 &  & InfoBERT & 95.6 & 39.0 & 59.21 & 95.6 & 44.0 & 53.97 & 95.6 & 34.0 & 64.44 & 95.6 & 18.6 & 80.54 \\
 &  & FreeLB++ & 95.4 & 34.8 & 63.52 & 95.4 & 36.2 & 62.05 & 95.4 & 23.8 & 75.05 & 95.4 & 10.4 & 89.1 \\
 &  & DSRM & 94.4 & 30.2 & 68.01 & 94.4 & 42.4 & 55.08 & 94.4 & 24.8 & 73.73 & 94.4 & 14.4 & 84.75 \\ \cline{3-15} 
 &  & PWWS + MMD (SemRoDe) & 93.8 & \textbf{50.0} & 46.7 & 93.8 & \textbf{59.6} & 36.46 & 93.8 & \textbf{50.2} & 46.48 & 93.8 & \textbf{34.8} & 62.9 \\
 &  & BERTAttack + MMD (SemRoDe) & 93.8 & (47.2) & 49.68 & 93.8 & {[}53.2{]} & 43.28 & 93.8 & 40.6 & 56.72 & 93.8 & 27.6 & 70.58 \\
 &  & TextBugger + MMD (SemRoDe) & 93.8 & 45.6 & 51.39 & 93.8 & {[}53.2{]} & 43.28 & 93.8 & {[}40.8{]} & 56.5 & 93.8 & {[}32.2{]} & 65.67 \\
 &  & TextFooler + MMD (SemRoDe) & 94.2 & {[}47.0{]} & 50.11 & 94.2 & (59.2) & 37.15 & 94.2 & (46.6) & 50.53 & 94.2 & (32.4) & 65.61 \\ \cline{2-15} 
 & \multirow{11}{*}{\textbf{MR}} & Vanilla & 89.6 & 24.0 & 73.21 & 89.6 & 33.2 & 62.95 & 89.6 & 12.0 & 86.61 & 89.6 & 5.0 & 94.42 \\
 &  & TextFooler + Adv Aug & 90.0 & 30.4 & 66.22 & 90.0 & 36.0 & 60.0 & 90.0 & 21.4 & 76.22 & 90.0 & 9.0 & 90.0 \\
 &  & TextFooler + Adv Reg & 90.8 & 31.8 & 64.98 & 90.8 & 34.6 & 61.89 & 90.8 & 21.0 & 76.87 & 90.8 & 9.4 & 89.65 \\
 &  & A2T & 92.0 & 24.2 & 73.7 & 92.0 & 28.8 & 68.7 & 92.0 & 15.0 & 83.7 & 92.0 & 4.4 & 95.22 \\
 &  & InfoBERT & 89.2 & 30.4 & 65.92 & 89.2 & 29.4 & 67.04 & 89.2 & 19.4 & 78.25 & 89.2 & 4.4 & 95.07 \\
 &  & FreeLB++ & 91.6 & 29.4 & 67.9 & 91.6 & 33.8 & 63.1 & 91.6 & 20.0 & 78.17 & 91.6 & 6.6 & 92.79 \\
 &  & DSRM & 88.4 & 29.0 & 67.19 & 88.4 & 39.8 & 54.98 & 88.4 & 25.2 & 71.49 & 88.4 & 9.8 & 88.91 \\ \cline{3-15} 
 &  & PWWS + MMD (SemRoDe) & 89.4 & {[}53.2{]} & 40.49 & 89.4 & \textbf{57.8} & 35.35 & 89.4 & (55.4) & 38.03 & 89.4 & (43.2) & 51.68 \\
 &  & BERTAttack + MMD (SemRoDe) & 89.8 & \textbf{53.6} & 40.31 & 89.8 & {[}56.8{]} & 36.75 & 89.8 & 46.2 & 48.55 & 89.8 & \textbf{56.4} & 37.19 \\
 &  & TextBugger + MMD (SemRoDe) & 89.4 & (53.4) & 40.27 & 89.4 & (57.2) & 36.02 & 89.4 & \textbf{55.6} & 37.81 & 89.4 & {[}42.8{]} & 52.13 \\
 &  & TextFooler + MMD (SemRoDe) & 88.8 & 51.8 & 41.67 & 88.8 & 56.6 & 36.26 & 88.8 & {[}54.8{]} & 38.29 & 88.8 & 42.2 & 52.48 \\ \hline
\end{tabular}
}\caption{
    Extended results: Values in \textbf{Bold} represent the highest scores, those in round brackets ($*$) denote the second highest, and values in square brackets [$*$] indicate the third highest.
    }\label{Appendix:distribution_alignment_mmd}
\end{table*}

\end{document}